\documentclass{article} %
\usepackage{iclr2027_conference,times}

\usepackage{amsmath,amsfonts,bm}

\def\eqref#1{equation~\ref{#1}}

\def\1{\bm{1}}

\DeclareMathAlphabet{\mathsfit}{\encodingdefault}{\sfdefault}{m}{sl}
\SetMathAlphabet{\mathsfit}{bold}{\encodingdefault}{\sfdefault}{bx}{n}

\usepackage{hyperref}
\usepackage{url}
\usepackage{tikz} %
\usepackage{wrapfig} %
\usepackage[nameinlink,capitalize]{cleveref} %
\usepackage{enumitem} %
\usepackage{booktabs} %
\usepackage{float} %
\usepackage{multirow}
\usepackage{algorithm}     %
\usepackage{algpseudocode} %
\usepackage{longtable}
\usepackage[breakable,skins,listings]{tcolorbox}
\usepackage{fontawesome5}

\definecolor{linkblue}{RGB}{32, 113, 190}

\title{Language Models that Play Chess \\and Explain Their Moves}

\author{Adithya Bhaskar\thanks{These authors contributed equally to this work.}\qquad Jeffrey Cheng\footnotemark[1] \qquad Danqi Chen\\
Princeton Language and Intelligence, Princeton University\\
\texttt{\{adithyab, jc93\}@cs.princeton.edu}}

\definecolor{claudeblue}{rgb}{0.10, 0.45, 0.85}

\newcommand{\ours}[0]{\textsc{Queen}}
\newcommand{\intermediate}[0]{\textsc{Pawn}}

\definecolor{darkblue}{rgb}{0, 0, 0.5}
\hypersetup{colorlinks=true, citecolor=darkblue, linkcolor=darkblue, urlcolor=darkblue}

\newcommand{\paperlink}[3]{%
  \href{#1}{\textcolor{darkblue}{#2\,\textbf{#3}}}%
}

\iclrfinalcopy %
\begin{document}

\maketitle

\begin{center}
\vspace{-2em}
\paperlink{https://huggingface.co/collections/princeton-nlp/queen-chess-models}{\faBrain}{Models}
\qquad
\paperlink{https://github.com/queen-project/queen}{\faGithub}{Code}
\qquad
\paperlink{https://queen-project.github.io/}{\faGlobe}{Website}
\end{center}

\begin{abstract}
Modern chess engines are silent experts: they play at a superhuman level, but do not offer explanations for their play.
On the other hand, language models (LMs) can generate plausible-sounding explanations, but their weak playing strength limits the utility of their explanations. We introduce \ours{}, a 4B-parameter chess-language model that can explain its moves and plans while playing at the level of a typical Grandmaster.
Our novel framework enables domain-specific reasoning through complementary components: an encoder-decoder architecture and an iterative distillation algorithm.
This architecture integrates a silent expert chess encoder with an instruction-tuned LM through cross-attention, which we train via a question-answering curriculum to extract chess concepts from the encoder's representations.
Building on this domain-adapted model, we iteratively improve its explanations with a natural-language analog of the Bellman update: the model analyzes the positions after its top candidate moves and consolidates them into an explanation of the current position, which is then distilled back into the model.
Over seven iterations, our model gains over 900 Elo points (1782 $\rightarrow$ 2697), substantially surpassing all frontier models on both playing strength and puzzle accuracy, despite containing three orders of magnitude fewer parameters.
Furthermore, LM-based evaluations show that our explanations are fluent and approach GPT-5.6-Sol (high) in coherence. The generality of our architecture and training procedure suggests a recipe for applying language models to domains where silent expert encoders are available, like games, robotics, and computer use.

\end{abstract}

\section{Introduction}

\begin{wrapfigure}{r}{0.4\textwidth}
    \centering
    \small
    \vspace{-4em}
    \includegraphics[width=0.35\textwidth]{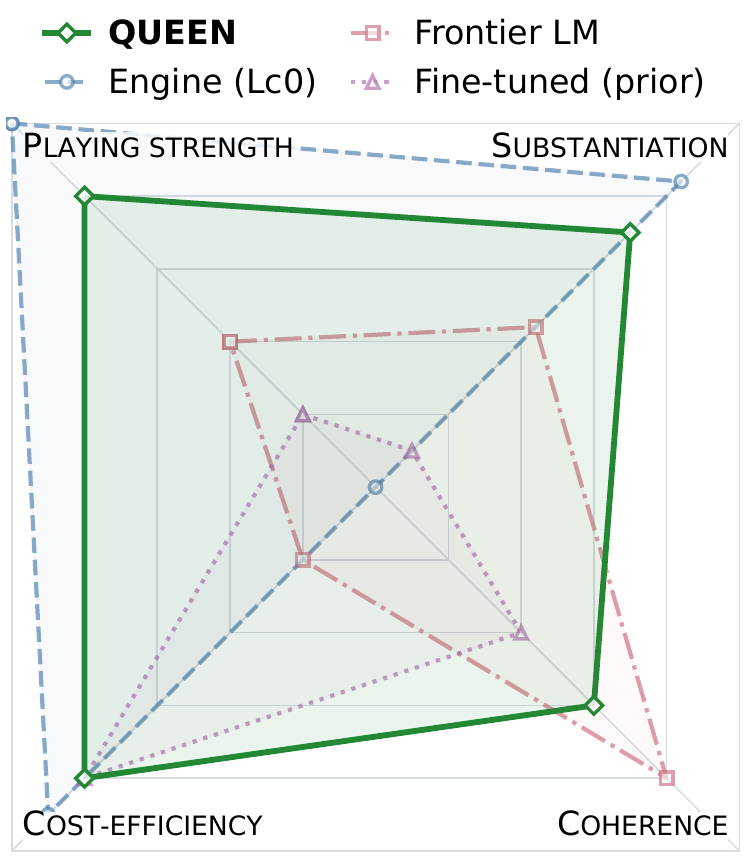}
    \vspace{-1em}
    \caption{Chess engines play strong moves but can't explain them. Frontier/finetuned LMs give plausible explanations, but are limited by weak play. \ours{} offers the best of both worlds.%
    \vspace{-2em}
    }
    \label{fig:example}
\end{wrapfigure}

In 1950, Claude Shannon estimated the computational complexity of chess and theorized what a chess-playing machine would look like. Chess became a benchmark for machine intelligence in the following decades, sparking a continual pursuit of chess programs stronger than the last \citep{Berliner1978}. This longstanding connection between chess and artificial intelligence has continued to evolve with advances in machine learning, most recently with the rise of Transformer models.

\begin{figure}[t]
    \small
    \centering
    \vspace{-1em}
    \includegraphics[width=0.97\linewidth]{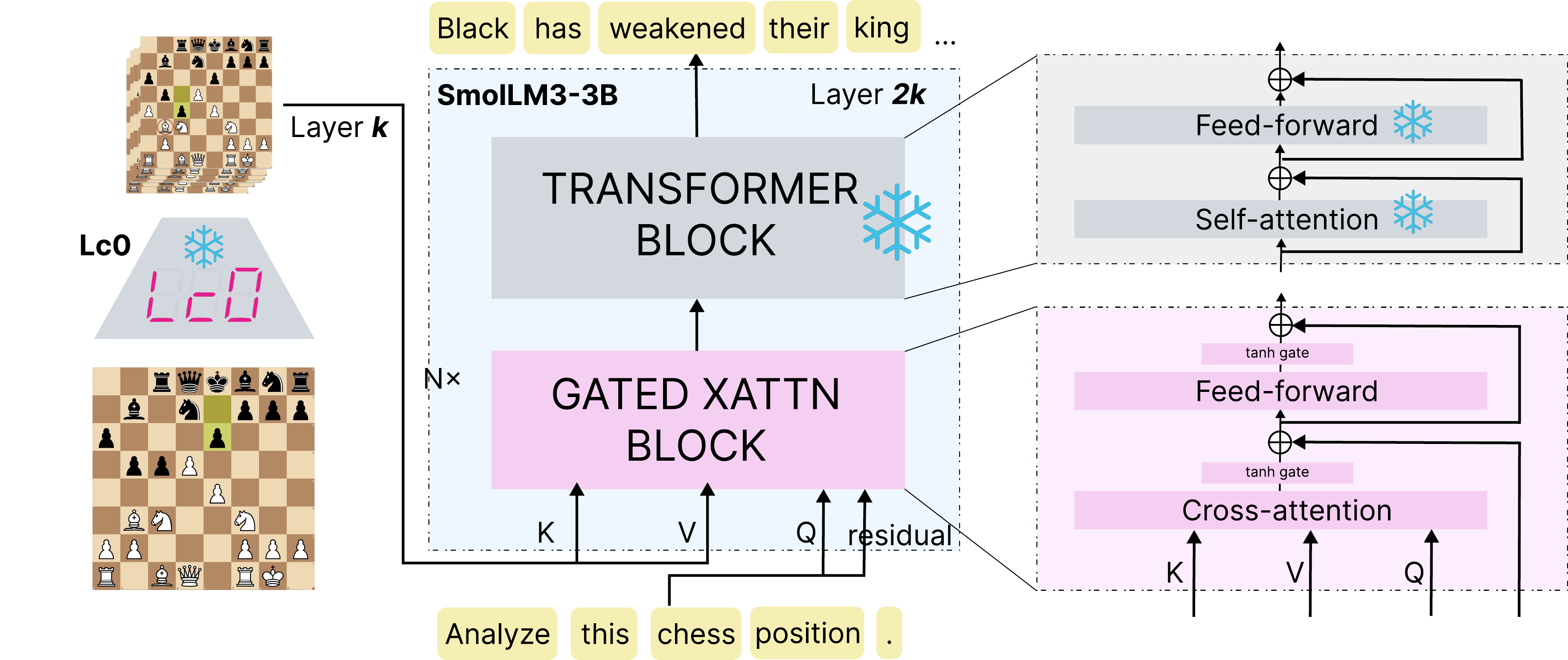}
    \vspace{-1em}
    \caption{We use an encoder-decoder architecture inspired by Flamingo~\citep{flamingo}. The encoder is an Lc0 chess network, and the decoder is an instruction-tuned language model. The $k$-th layer embeddings of Lc0 flow into a Flamingo block injected before the $2k$-th LM layer, and all previously trained parameters are frozen during domain adaptation.}
    \label{fig:arch}
    \vspace{-2.5em}
\end{figure}

We present \ours{} (\textbf{QU}ality \textbf{E}xplanation and \textbf{E}valuation \textbf{N}etwork), an encoder-decoder model capable of playing chess at 
a Grandmaster
level while providing high-quality explanations justifying its moves (\cref{fig:arch}).\footnote{As of October 2, 2026, there are 1,899 Grandmasters, fewer than 0.01\% of active online players.}  Good explanations are an accessible way for chess players to understand and learn from expert predictions; they can also serve as valuable post-training data for language models.
However, existing approaches fall short.
Modern chess engines are silent experts: Transformer-based chess engines achieved superhuman playing strength but don't communicate their reasoning in natural language \citep{googlechesstransformer, leelachesszero}. On the other hand, language models (LMs) have been trained to explain chess positions in natural language, but these models have largely been trained on and evaluated using isolated puzzle positions rather than full games. They often struggle to select strong moves even in this restricted setting, limiting the quality of the explanations they provide~\citep{cui2026communicatingchessstrategiesnatural, kim2025bridginggapexpertlanguage, tang2026groundedchessreasoninglanguage}. Today's frontier LMs (GPT-5.6-Sol, Gemini-3.1-Pro) have greatly improved over their predecessors \citep{kolasani2025llmchess} and can generate adequate explanations, but their substantial computational cost limits their practicality. 

In this work, we address this gap through two complementary components. We first propose an encoder-decoder architecture that melds a silent expert encoder with a standard decoder LM with a cross-attention mechanism. We then train the cross-attention parameters through a curated QA curriculum that 
progresses from static board-state understanding to dynamic move transitions in future positions. 
Next, we introduce an iterative distillation algorithm inspired by the Bellman value update, serving as a natural-language analogue to its real-valued counterpart  \citep{bellman1957markovian}.
At each iteration, we
combine the explanations and verdicts of child nodes to construct an explanation of the root position—including why the chosen move is best and why alternatives are inferior—which is then distilled back into the model. This procedure increases the quality of our model's explanations by improving its implicit search and look-ahead capabilities. 

To assess the quality of our model's explanations, we establish an evaluation framework along three dimensions: \textit{accuracy}, \textit{substantiation}, and \textit{coherence}. Our model's playing strength, a proxy for the \textit{accuracy} of its recommendations, reaches a final estimated rating of 2697, far surpassing frontier models such as GPT-5.6-Sol (2071) and Gemini-3.1-Pro (2201) and approaching the Lichess blitz rating of the median Grandmaster (2730). Its ability to \textit{substantiate} these recommendations is evaluated through tactical puzzles, where its no-mistake rate is higher than all baselines.
Finally, LM-based evaluation finds that its explanations are fluent and approach GPT-5.6-Sol in \textit{coherence}. 

Beyond chess, our framework suggests a general recipe for coupling language models with silent experts to produce high-quality natural-language explanations, with potential applications in games, robotics, and computer use. 
 The contributions of this paper are as follows:
\begin{itemize}[nosep]
    \item We propose an encoder-decoder architecture and subsequent training recipe to adapt LMs
    to the chess domain. While our experiments focus on chess, our pipeline is general and can be easily extended to settings where a domain-expert Transformer encoder can be trained. \vspace{1pt}
    \item We introduce an iterative search-distillation algorithm that improves the model's implicit search and look-ahead capabilities. This algorithm is general and can be extended to any domains that benefit from tree-search methods such as MCTS and alpha–beta pruning. \vspace{1pt}
    \item We establish an evaluation framework for assessing the quality of generated explanations along three dimensions: accuracy, substantiation, and coherence. %
    Using this framework, we systematically evaluate our model alongside prior work and frontier models.
\end{itemize}

\section{Related Works} 
\label{sec:relwork}
\paragraph{Chess engines.}
\label{sec:chessengine}

Chess engines are computer programs that play chess by exploring candidate moves and continuations through \textit{search} and using an \textit{evaluator} to assess the resulting positions in order to determine best continuations. 
The evaluator assigns each position a scalar, real-valued evaluation in \textit{centipawns}, which can be converted to expected win rates for either side.\footnote{A centipawn, 1/100 of a pawn, is the unit of measure used in chess to quantify the advantage of a player.} Chess engines also provide a sequence of best moves from both sides, known as a \textit{principal variation} (PV). At sufficiently high search budgets, modern engine evaluations and PVs can be treated as oracle. Chess engines mainly differ on how they generate evaluations and perform search. The first engine to defeat a world champion, Deep Blue, used a complex handcrafted evaluation (HCE) function combined with alpha-beta pruning and other chess heuristics \citep{CAMPBELL200257}. 
Then, AlphaZero demonstrated that a neural network trained purely through self-play, could surpass the strongest HCE-based engines \citep{alphazero}.
Traditional chess engines like Stockfish eventually shifted from HCEs in favor of neural network evaluations ~\citep{stockfish}.
The advent of Transformers led to their adoption in chess: 
Leela Chess Zero (Lc0) and Google DeepMind released Transformer-based chess engines capable of near-superhuman play~\citep{googlechesstransformer, monroe2026chessformerunifiedarchitecturechess}.
Another line of work instead builds human-aligned engines: Maia~\citep{McIlroy_Young_2020,tang2024maia2unifiedmodelhumanai} and Allie~\citep{zhang2025humanaligned} are trained to predict moves that human players of a given skill level would make, rather than the strongest move, enabling them to serve as human-like opponents and teaching tools. We share this goal of making chess AI useful to human players, but pursue it differently: whereas chess engines output only move predictions, \ours{} additionally explains the reasoning behind them in natural language.

\paragraph{Language models and chess.}
Prior work showed that autoregressive Transformers trained with next-token prediction on chess data can acquire substantial chess-playing ability~\citep{zhang2024transcendence, schultz2025mastering}. However, adapting general-purpose pretrained language models to chess proved substantially more difficult~\citep{noever2020chess, zhang2025completechessgamesenable, hwang2025largelanguagemodelsdevelop}. Consequently, benchmarks focused on narrow components of chess capability, progressing from board-state tracking~\citep{toshniwal2022chess, chessgpt} to move selection among candidates~\citep{mate} and high-level strategic motifs~\citep{chessqa}. In parallel, several lines of research sought to make chess expertise accessible. Early work focused on generating natural-language commentary for chess moves~\citep{jhamtani2018learning, zang2019automated, lee2022improving}, while related work extracted chess concepts encoded in the latent representations of chess engines such as AlphaZero~\citep{mcgrath2022acquisition, schut2025bridging} and Lc0~\citep{jenner2024evidencelearnedlookaheadchessplaying}. \citet{kim2025bridginggapexpertlanguage} bridged these directions by combining extracted chess concepts with language models to generate commentary, but remained conditioned on a given move. Recently, efforts instead shifted from generating commentary to explanations, where the model must first predict a move and then provide the reasoning behind its choice. This has been pursued by prompting~\citep{cui2026communicatingchessstrategiesnatural} and distilling~\citep{tang2026groundedchessreasoninglanguage} frontier language models. Concurrent work adopted an encoder-decoder architecture that integrated latent representations from a silent chess expert into a language model~\citep{s2026exploringcollaborationlanguagenonlanguage}, similar to our approach. The quality of an explanation depends on the quality of the move it explains, making chess-playing ability an important prerequisite for useful explanations. Compared to prior methods, \ours{} achieves a substantially greater playing strength, enabling it to generate higher-quality explanations.

\section{\ours{}: Our Approach}

Our goal is to generate natural language explanations of chess positions. A high-quality explanation should consist of three components (\cref{fig:format}): a best move recommendation, a predicted principal variation (PV), and natural-language prose explaining the two. The next move provides a concrete recommendation for the position, while the PV substantiates this recommendation by demonstrating the consequences of the predicted move. Finally, the prose makes this analysis accessible by articulating the reasoning underlying the predicted move, such as analyzing alternative moves.

There are two natural starting points to achieve this goal: silent Transformer-based chess engines capable of playing chess at superhuman levels, and general-purpose LMs able to generate fluent text but lacking chess expertise.
These chess engines are not suited to natural-language generation: 

\begin{wrapfigure}{r}{0.4\textwidth}
\centering
    \includegraphics[width=0.39\textwidth]{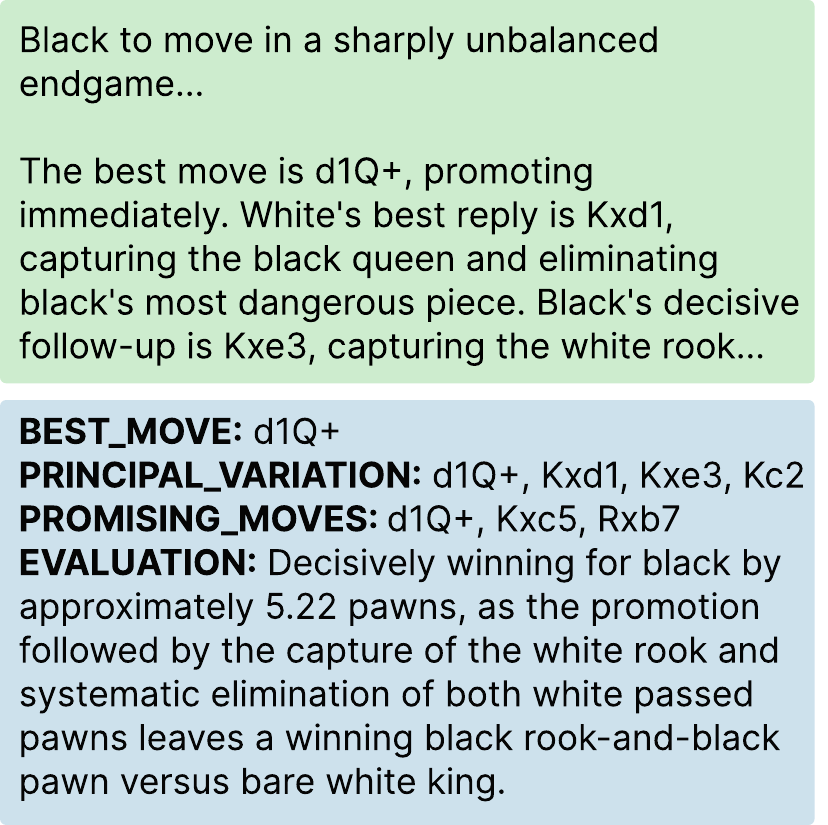}
    \vspace{-0.5em}
    \caption{An example explanation generated by \ours{}, snipped for readability. The explanation consists of natural language prose (green) followed by a best move and PV prediction (blue). %
    }
    \label{fig:format}
    \vspace{-1em}
\end{wrapfigure}

they operate at the hundred-million parameter scale and only produce a policy over legal moves and win/ draw/ loss probabilities. Extending them to generate natural-language would require substantial changes to their architecture and training objective.
In contrast, finetuning LMs to play chess has proven difficult; existing approaches achieve only modest playing strength~\citep{zhang2025completechessgamesenable,hwang2025largelanguagemodelsdevelop}. Giving explanations grounded in strong play presents a greater challenge.

With these considerations in mind, we propose an encoder-decoder architecture consisting of a silent chess encoder and a pretrained LM decoder.
Our insight is that the LM must reason over chess concepts like positional features and tactical motifs, and how these concepts change under candidate moves to produce good explanations.
Rather than expecting the LM to acquire these concepts and domain-specific reasoning ability from expert traces, which typically consist of state-action pairs without natural language explanations, we instead bridge the LM directly to a model whose representations encode these concepts~\citep{jenner2024evidencelearnedlookaheadchessplaying}.
We provide an in-depth description of our architecture in \cref{sec:arch}.
This design decision yields a natural two-stage training process: a domain adaptation phase where we train the bridge parameters to interpret the encoder hidden states (\cref{sec:adapt}); and the main training phase where the model learns to produce high-quality explanations (\cref{sec:improve}).

\subsection{Architecture}
\label{sec:arch}

\paragraph{Chess encoder.}
We choose the BT5 network from the Lc0 family~\citep{monroe2026chessformerunifiedarchitecturechess}, henceforth referred to as Leela. Leela is a 240M parameter encoder-only model, consisting of 15 layers with a hidden dimension of 1024. It always operates on a sequence length of 64 tokens, each corresponding to a square on the chessboard, and plays at a near-superhuman level without search.

\vspace{-0.75em}

\paragraph{Language model decoder.} We use SmolLM3-3B model as our LM decoder \citep{bakouch2025smollm3}. SmolLM3 is instruction tuned, but is a general-purpose language model with little to no chess-specific training. It has 36 layers with a hidden dimension of 2048. 

\vspace{-0.75em}

\paragraph{Bridge architecture.} 
Encoder-decoder architectures that pair domain-specific encoders with decoder LMs have been extensively explored through vision-language models. These approaches introduce mechanisms to project the visual encoder features into representations that can be processed by the decoder. We take inspiration from one such approach, Flamingo~\citep{flamingo}, which connects the encoder and decoder with cross-attention. Let $e_i$ and $d_i$ denote the output of the $i$th encoder and decoder layers, respectively.
We insert \textit{gated cross-attention} layers operating on a pair $(e_i, d_j)$ between the usual decoder layers, where key and value projections come from $e_i$ and query projections come from $d_j$.
Unlike the original Flamingo architecture where only the last encoder hidden state is used, we pair $e_i$ with $d_{2i}$, and insert the corresponding gated cross-attention layer prior to the $2i$th decoder block. This choice allows the decoder to access representations from earlier stages of the encoder, rather than relying solely on its final hidden state.

\vspace{-0.75em}

\paragraph{Full model flow.} Inputs to the model are a pair of (FEN, text).\footnote{Any chess position can be succinctly encoded using \href{https://en.wikipedia.org/wiki/Forsyth-Edwards_Notation}{Forsyth-Edwards Notation} (FEN).} The FEN string is processed by Leela to produce a sequence of encoder hidden states,
which are provided as context to the SmolLM decoder through the gated cross-attention layers. Thus, the decoder can condition on the chess position at every position in the text sequence while autoregressively generating the output. We provide a diagram of the overall architecture in Figure~\ref{fig:arch}.

We ablate both our design choices for the language model decoder and the bridge architecture. On the decoder side, we try with Qwen and Gemma models of similar size (4B parameters). For the architectural ablations, we experiment with passing the projected encoder hidden states directly to decoder key-value cache as well as a LLaVA-inspired architecture~\citep{llava}. The results of these ablations and additional architecture details can be found in~\cref{app:domain}.

\begin{figure}[H]
    \small
    \centering
    \includegraphics[width=0.98\linewidth]{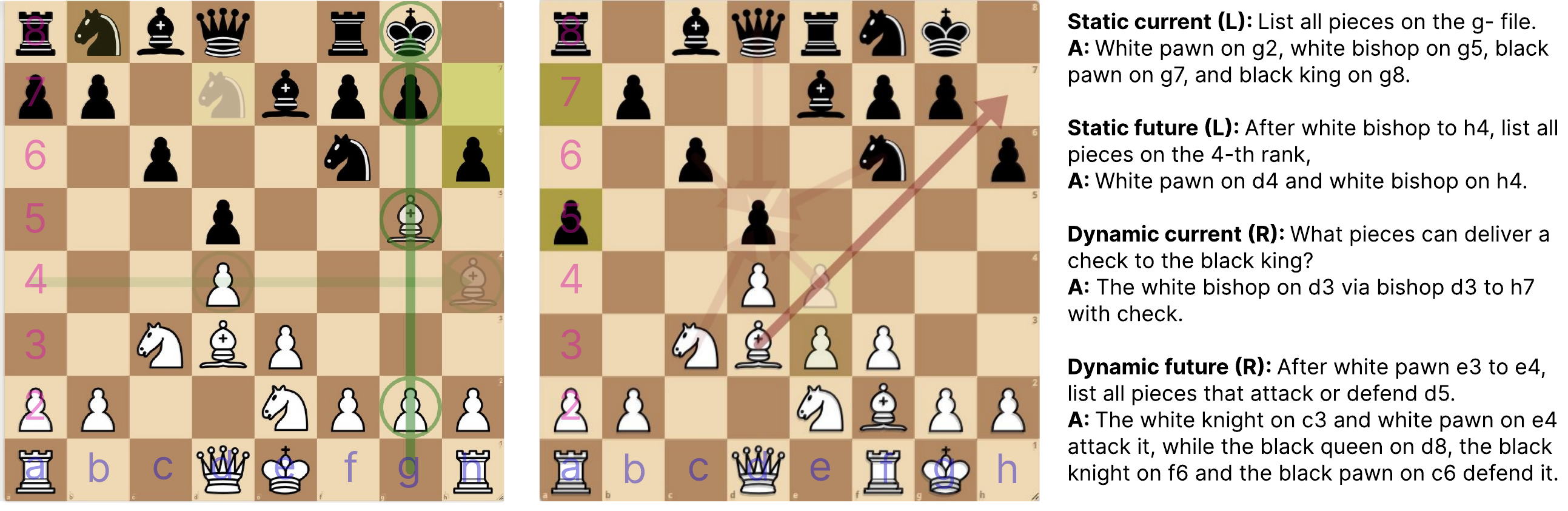}
    \caption{
    Domain adaptation progressively teaches the model to extract static features in the current board state (\textsc{static-current}), reason about available moves and interactions (\textsc{dynamic-current}), reconstruct a future board state from move sequences (\textsc{static-future}), and finally reason about the resulting position (\textsc{dynamic-future}). 
    }
    \label{fig:curriculum}
    \vspace{-1em}
\end{figure}

\subsection{Domain Adaptation: Interpreting Encoder Hidden States}
\label{sec:adapt}

Prior work showed that Leela encodes aspects of chess positions such as piece arrangement, available legal moves, and tactical continuations \citep{jenner2024evidencelearnedlookaheadchessplaying}. Thus, the goal of the domain adaptation phase is to train the decoder to reliably extract these latent features from Leela's representations. To this end, we curate a question-answer dataset where each example consists of a position $p$, represented by its FEN encoding, and a query $q$. Every query $q$ is asked about the current encoded position $p$ or a future position $p'$ reached by a provided sequence of 1 to 8 moves. The positions $p$ are sampled from the publicly available Lichess dataset \citep{lichess_database} and the queries $q$ fall into four categories, illustrated in \cref{fig:curriculum} and below. Each question type contains three query and answer formats to promote diversity, and all answers are generated and verified programmatically.
\begin{enumerate}[nosep]
    \item \textbf{\textsc{Static-Current}}: These questions teach the model to recover basic board-state information from Leela's representation of the current position, such as identifying the piece on a queried square, finding the location of a queried piece, or listing all pieces.

    \item \textbf{\textsc{Dynamic-Current}}: These questions teach the model to extract rules about how pieces move and interact in the current position, such as finding all possible legal moves for a queried piece or listing all possible captures and checks.

    \item \textbf{\textsc{Static-Future}}: These questions require the model to first reconstruct the position resulting from a supplied sequence of moves and then answer a static query about the resulting board, such as identifying pieces on particular squares or ranks.

    \item \textbf{\textsc{Dynamic-Future}}: These questions combine future-state reconstruction with reasoning about the resulting position, such as finding legal moves, captures, checks, or attackers after a supplied sequence of moves.
\end{enumerate}
Answering questions about chess positions requires reliably representing piece identities and board squares. Because the tokenizer inconsistently represents square names and pretrained embeddings may encode undesirable linguistic priors for chess piece names (e.g., the non-chess meanings of “bishop” and “knight”), we add dedicated tokens for all 64 squares and 12 piece types.

We construct a separate dataset for each question type and perform domain adaptation sequentially, progressing from \textsc{Static-Current} $\rightarrow$ \textsc {Dynamic-Current} $\rightarrow$ \textsc{Static-Future} $\rightarrow$ \textsc{Dynamic-Future}. This progression teaches the model to first extract all features of the encoded board state; future-position stages introduce an additional state-tracking challenge: the model must first reconstruct the board state resulting from the provided move sequence before answering the query. Question types from earlier stages are replayed at low proportions in subsequent stages to prevent forgetting. 
In all stages, the encoder and decoder are both frozen; only the cross-attention bridge parameters and new token embeddings are trained. 
We use the AdamW optimizer with a cosine learning-rate schedule. 
All stages use early stopping based on the validation set, and the best checkpoint from each stage is used as the initialization for the next.
We denote the model checkpoint obtained after this training as \intermediate{} (\textbf{P}osition \textbf{AW}are \textbf{N}etwork). Additional details on this stage, the exhaustive list of questions and training hyperparameters are presented in
\cref{app:domain}.

\newpage
\subsection{Iterative Search Distillation: Improving Explanation Quality}
\label{sec:improve}

After domain adaptation, \intermediate{} can answer structured chess questions, but has not seen examples of our desired explanation format. We therefore seed the model with examples of this form generated by GPT-5.6-Sol (low). Concretely, we source 15,000 positions from Lichess games and puzzles and prompt Sol to produce explanations consisting of a best move prediction, PV prediction, and natural-language prose, together with three promising moves and a position evaluation (\cref{fig:format}). The latter two fields are also used to guide the search process in our iterative distillation algorithm. We perform SFT after filtering out responses containing illegal moves or mistakes, yielding \intermediate-1.

\begin{figure}[t]
    \centering
    \includegraphics[width=0.98\linewidth]{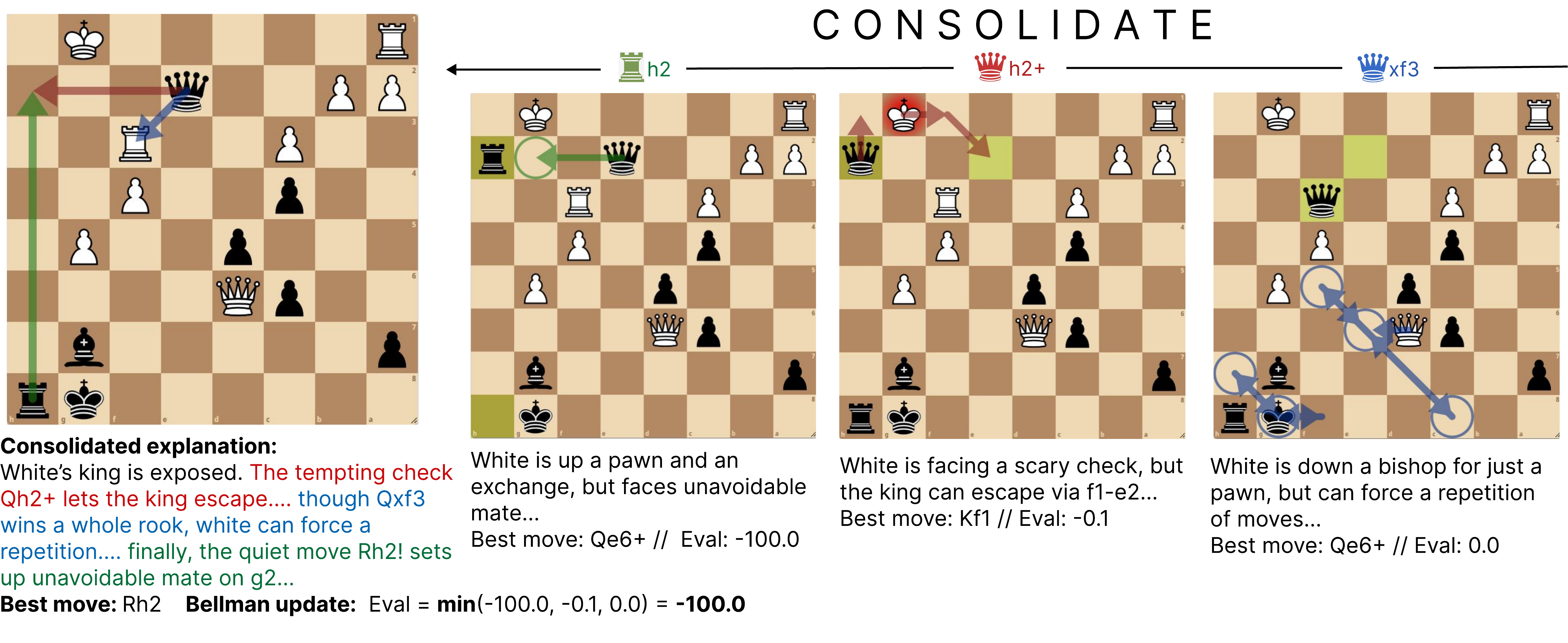}
    \vspace*{-1em}
    \caption{We repeatedly distill the search-consolidated explanations back into the model for iterative self-improvement. Analogous to the Bellman update, a consolidated explanation combines the explanations at child nodes to explain why one move is the best and the shortcomings of other moves.}
    \label{fig:selfdistill}
    \vspace{-1em}
\end{figure}

Though fluent, \intermediate-1's explanations recommend low-quality and illegal moves, likely due to the limited strength of the teacher and the small size of the seed dataset.
To improve the quality of its explanations, we look to AlphaZero \citep{alphazero} for inspiration: its paradigm repeatedly uses MCTS to produce improved evaluations, which are distilled into the network so that it can reproduce them without search.
Replacing MCTS with alpha-beta search gives a simplified form of Bellman value iteration~\citep{bellman1957markovian}, an approach also used by~\citet{schultz2025mastering} to train a Transformer to act as an action-value network. We extend this idea to natural language, proposing an analogue of the Bellman update where improved explanations are distilled back to the model (\cref{fig:selfdistill}). In practice, we apply the below process to train \intermediate-$(k+1)$ from \intermediate-$k$.

\begin{enumerate}[nosep]
\item \textbf{\textsc{Sample}}: We sample a pool of roughly 400K root positions from three sources: (a) games played between the current checkpoint and Stockfish at different node counts, (b) games played between humans on Lichess, and (c) puzzle positions mined from Lichess puzzles.

\item \textbf{\textsc{Generate}}: For each root position $p$, \intermediate-$k$ generates an explanation containing three promising moves. If all three moves are mistakes according to Stockfish,\footnote{A \textit{mistake} is a move that incurs a 10\% or greater drop in expected win rate compared to Stockfish oracle.} we replace the worst move with the Stockfish oracle move. \intermediate-$k$ then independently generates explanations of the three resulting positions. 

\item \textbf{\textsc{Recurse}}: We inspect the best-move prediction from each child explanation. If any child predicts a mistake, we discard the current root and its other children, and repeat the prior step from the selected child. We recurse until the next-move predictions of all three child explanations are mistake-free or the maximum recursion depth is reached. 

\item \textbf{\textsc{Consolidate}}: We combine the three child explanations using Qwen3.8-27B (instructed not to introduce any new content) to produce a consolidated explanation of the root position.

\item \textbf{\textsc{Train}}: We filter out instances by comparing the PVs in the consolidated and original explanations at their first divergence, discarding the consolidated explanation if its move has a worse Stockfish evaluation or is illegal. The remaining explanations form fine-tuning data for \intermediate-$(k+1)$, which we obtain by SFT from \intermediate-$k$.

\end{enumerate}

We run seven iterations of training, with \intermediate-8 promoted to the name \ours.
Full details and hyperparameters of the initial seeding and iterative training phases are provided in \cref{app:iterative}.

\newpage
\section{Evaluations}
\label{sec:eval}

We identify three desiderata for high-quality explanations. First, they should be \textbf{accurate}: the recommended move should be near-optimal; an explanation is not useful if it recommends a blunder. Second, an explanation should \textbf{substantiate} its recommended move by analyzing alternate variations. A good move recommendation alone does not establish why the move is preferable to its alternatives; an accurate PV provides evidence that the model can reason about the consequences that distinguish similar candidate moves. Finally, explanations should be \textbf{coherent}, containing fluent prose, referencing high-level strategic and tactical motifs, and having few hallucinations.

We assess accuracy, measured by the quality of the model's move recommendations, through full-game simulations against fixed opponents, reporting estimated Elo ratings to quantify playing strength (\cref{sec:accuracy}). 
Next, we assess our explanations' substantiation on a held-out set of positions to see whether the model can produce mistake-free principal variations (\cref{sec:substantiation}).
Finally, we assess their coherence using an LM-based judge (\cref{sec:coherence}).

For our baselines, we compare our model against three frontier language models, GPT-5.6-Sol (\textsc{Sol}), GPT-5.6-Luna (\textsc{Luna}), and Gemini-3.1-Pro (\textsc{Gemini}), under \textit{high reasoning} budgets. We also evaluate C1-4B~\citep{tang2026groundedchessreasoninglanguage}, a similarly sized model from prior work trained through chess-specific master distillation.\footnote{We would also like to compare against LLAMIA~\citep{s2026exploringcollaborationlanguagenonlanguage}, concurrent work that adopts a similar architecture; however, its models have not been released as of this writing.}
Full details for all evaluations are reported in \cref{app:evals}.

\subsection{Accuracy on Full Game Simulations}
\label{sec:accuracy}

\begin{wraptable}{r}{0.4\textwidth}
    \centering
    \small
    \vspace{-1em}
    \begin{tabular}{lc}
        \toprule
        \textbf{Model} & \textbf{Elo} \\
        \midrule
        \ours{} \textbf{(ours)} & \textbf{2697} \\
        \textsc{Gemini} & 2201 \\
        \textsc{Sol} & 2071 \\
        \textsc{Luna} & 1822 \\
        C1-4B & 514* \\
        \midrule
        Median Grandmaster & 2730 \\
        Median International Master & 2560 \\
        Median FIDE Master & 2470 \\
        \midrule
        \textit{Leela (BT5 Encoder)} & \textit{2987} \\
        \bottomrule
    \end{tabular}
    \caption{Estimated ratings anchored to Lichess Elos after 32 games. All frontier models use high reasoning effort. \\
    \textit{*C1-4B lost all 32 games.}}
    \label{tab:elo}
    \vspace{-1em}
\end{wraptable}

The primary purpose of performing full game simulations is to evaluate the \textit{accuracy} of our model's explanations. We believe that evaluating on full games authentically simulates real-world human-model interactions and offers greater robustness compared to just measuring best move accuracy on a static evaluation set~\citep{zhang2025completechessgamesenable}. We choose a list of 8 chess engines of varying strengths and pit each evaluated model against the 8 engines. Each model plays a total of 32 games, and we report the resulting Elo scores from their performance in \cref{tab:elo}, anchored with the known strength of the Leela networks to the Lichess Elo scale.

Overall, \ours{} achieves the highest rating out of all evaluated models, beating \textsc{Sol} by over 600 rating points and \textsc{Gemini} by over 450. The magnitude of these differences is substantial, corresponding to an expected 97.4\% and 94.6\% win rate, respectively. We attribute this strong performance to our iterative search distillation procedure: across seven iterations, \ours{} gains 915 Elo points, (1782 $\rightarrow$ 2024 $\rightarrow$ 2187 $\rightarrow$ 2346 $\rightarrow$ 2539 $\rightarrow$ 2434 $\rightarrow$ 2559 $\rightarrow$ 2697). The gap between \ours{} and Leela reflects the challenge of expressing latent expert knowledge in language, which we view as a form of \textit{verbalization debt}~\citep{s2026exploringcollaborationlanguagenonlanguage}. Nonetheless, \ours{} retains high accuracy in its move recommendations, providing a foundation for generating high-quality explanations.

\subsection{Substantiation on Held Out Positions}
\label{sec:substantiation}

We measure \textit{substantiation} by evaluating the moves in the predicted principal variation (PV). We report the \textbf{no-mistake rate} (NMR), the fraction of predicted PVs containing no mistakes, and the \textbf{first-move no-mistake rate} (FNMR), the fraction of positions in which the model's first move is not a mistake. The latter serves as a proxy for move accuracy, allowing comparison with prior work that cannot generate a PV without re-encoding the position at each step.

We first evaluate on 1,000 tactical puzzles spanning projected difficulties of 609 through 2905 on Lichess. 
The puzzles present positions where a unique sequence of moves leads to an advantage, often involving temporary concessions like sacrificing a piece. Thus the entire sequence must be analyzed without error to adequately explain the position.
Since tactical puzzles only capture a narrow proportion of chess positions, we additionally measure the same two metrics on a set of 1,000 general positions (drawn from human games) to supplement the former set. 

\begin{table}[t]
\centering
\small
\begin{tabular}{l c c c c}
\toprule
\multirow{2}{*}{\textbf{Model}} & \multicolumn{2}{c}{\textbf{Tactical} $\uparrow$} & \multicolumn{2}{c}{\textbf{General} $\uparrow$} \\
 \cmidrule{2-3}  \cmidrule{4-5} 
& \textbf{NMR (\%)} & \textbf{FNMR (\%)} & \textbf{NMR (\%)} & \textbf{FNMR (\%)}\\
\midrule
\ours & \textbf{68.8} & \textbf{91.6} & \textbf{66.6} & \textbf{97.1} \\
\textsc{Gemini} & 66.9 & 82.5 & 63.0 & 89.0 \\
\textsc{Sol} & 65.1 & 81.9 & 61.5 & 86.8 \\
\textsc{Luna} & 35.6 & 53.5 & 47.1 & 73.6 \\
C1-4B & --- & 43.2 & --- & 38.8 \\
\bottomrule
\end{tabular}
\caption{Substantiation evaluation on 1,000 tactical and 1,000 general positions. \textbf{NMR} is the fraction of predicted PVs containing no mistakes; \textbf{FNMR} is the fraction of positions in which the model's predicted best move is not a mistake. All frontier models use high reasoning effort. \ours{} performs the best across all metrics on both tactical and general positions.}
\vspace{-1em}
\label{tab:positions}
\end{table}

The results are shown in \cref{tab:positions}. 
\ours{} performs the best across all metrics, beating out \textsc{Gemini} by 2 points on NMR, and 9 points on FNMR.
It beats other models by an even larger margin, including the model it was seeded from (\textsc{Sol}). 
C1-4B, the non-frontier baseline, ranks far below other models.
Since \ours{} finds good moves much more frequently than other methods and also substantiates the critical continuation better, its explanations are more informative as a whole.

\subsection{Coherence under Frontier LM Judging}
\label{sec:coherence}

While the previous two metrics evaluated the next move recommendation and PVs contained in the explanation, \emph{coherence} evaluates its prose. We distinguish between three notions of coherence. \emph{Structural coherence} evaluates the implicit tree of discussed moves: whether reasonable alternatives and critical variations are considered, independent of how they are described. Because the relevant variations depend on the recommended move, structural coherence is orthogonal to accuracy and substantiation. \emph{Conceptual coherence} evaluates whether the language appropriately uses high-level concepts and motifs (e.g. forks, pins, outposts), is free of hallucinations, and accurately captures each move's intentions and consequences. Finally, \emph{fluency} evaluates the linguistic quality of the explanation, including grammaticality, clarity, and readability. We evaluate these three axes through annotation by a GPT-5.6-Sol (high) judge on a Likert scale (from 1 to 5).

\begin{table}[H]
    \centering
    \small
    \begin{tabular}{l c c c }
    \toprule
    \textbf{Model} & \textbf{Structural Coherence} & \textbf{Conceptual Coherence} & \textbf{Fluency} \\
    \midrule 
    \ours & 3.51 & 2.76 & 4.45 \\
    \textsc{Gemini} & \textbf{3.55} & 3.30 & 4.98 \\
    \textsc{Sol} & 3.51 & \textbf{3.61} & \textbf{4.99} \\
    \textsc{Luna} & 2.82 & 2.94 & 4.96 \\
    C1-4B & 2.48 & 2.16 & 4.94\\
    \bottomrule
    \end{tabular}
    \caption{Evaluation of the various models on coherence and fluency. Frontier models use high reasoning effort. All models are fluent in their output. \textsc{Gemini} obtains the best structural coherence, though \ours{} and \textsc{Sol} obtain comparable scores. \ours{} struggles with conceptual coherence, as iterative search-distillation cannot teach it to verbalize and describe motifs it hasn't seen.}
    \label{tab:explanation}
    \vspace{-1em}
\end{table}

We find in Table~\ref{tab:explanation} that all models are fluent. 
\ours{} achieves a structural coherence score of 3.51, tied with \textsc{Sol} and approaching \textsc{Gemini}, implying that it finds illustrative lines comparably well to the two models.
One reason the score is not even higher for \ours{} is that it considers an average of 30.0 ply in the entire analysis, while \textsc{Sol} only considers 23.8 -- giving the judge a greater chance to penalize it.
We also note that \ours{} obtains a low score of 2.76 on conceptual coherence. 
While repeated self-improvement gradually distills an analysis tree into model weights (and thus lifts structural coherence), hallucinations of motifs and patterns can potentially propagate through the consolidation and keep conceptual coherence low.
This is a limitation of our method and we look to future work to resolve it.
We conclude that \ours{} provides fluent, high-quality explanations.

\newpage
\section{Analysis}

\paragraph{Ablations.} 
\begin{wraptable}{r}{0.45\textwidth}
\small
\centering
\vspace{-1em}
\begin{tabular}{lrr}
\toprule
\textbf{Model} & \textbf{Elo} $\uparrow$ & \textbf{Tactics} $\uparrow$ \\
\midrule
 \intermediate-1 & 1782 & 73.9 \\
\quad $-$ Encoder & 514$^*$ & 9.7 \\
\quad $-$ Curriculum & 514$^*$ & 1.3 \\
\bottomrule
\end{tabular}
\caption{Ablations of \ours{} evaluated after \textsc{Sol}-seeding. Both the encoder and curriculum are essential for a good initialization. 
}
\label{tab:ablations}
\vspace{-1em}
\end{wraptable}
We ablate the Leela encoder and domain-adaptation stage from \ours{} to quantify their importance. Because iterative search distillation requires costly data generation, we evaluate the ablations immediately after the GPT-5.6-Sol seeding stage.
The encoder ablation replaces Leela with a dictionary-form representation of the complete board state (e.g., ``White King: e1, ...''), containing the same information as the FEN. The representation is passed directly to the LM without an encoder or cross-attention, while retaining the same domain-adaptation curriculum.
The curriculum ablation skips the domain adaptation phase of training but keeps the Leela encoder. 
We evaluate both on playing strength Elo and first-move no-mistake rate (FNMR) on the 1,000 tactical positions. As shown in Table~\ref{tab:ablations}, removing either component substantially degrades the model, indicating that both are important for producing a strong initialization for iterative search-distillation.

\begin{wraptable}{l}{0.35\textwidth}
\vspace{-1em}
\centering
\small
\begin{tabular*}{\linewidth}{lc}
  \toprule
  \textbf{Outcome} & \textbf{Rate (\%)} \\
  \midrule
  Agrees with Lc0 & 46.2 \\ 
  Disagrees, in top-5 & 39.1 \\ 
  Outside top-5 & 14.7  \\
  \bottomrule
\end{tabular*}
\caption{\ours{} disagrees with Leela on 53.8\% of positions with five or more reasonable moves, suggesting that it does not simply reproduce Leela's policy head.}
\label{tab:agreement}
\vspace{-2em}
\end{wraptable}
\paragraph{(Dis)agreement with Leela.}

A shortcut we'd like to guard against is the model learning to reconstruct Leela's policy head and then attaching a post-hoc explanation to its recommended move. We investigate this on a set of 1,000 positions where at least five moves have expected win rates within 3\% of the Stockfish-optimal move. The choice among several near-optimal moves more directly reflects the model's preferences. \cref{tab:agreement} shows that on 538 of the 1,000 positions, our model selects a move different from Leela's choice. This suggests that \ours{} does not simply reproduce Leela's policy, but instead uses the Lc0 representations to inform its own move selection.

\begin{wraptable}{r}{0.4\textwidth}
\centering
\scriptsize
\vspace{-1.5em}
\begin{tabular}{lrrrr}
\toprule
\multirow{2}{*}{\textbf{Model}} & \multicolumn{4}{c}{\textbf{Elo} $\uparrow$}\\
\cmidrule{2-5}
& P1 & P2 & P3 & P4\\
\midrule
\ours{} & 1782 & 2024 & 2187 & 2346 \\
\ours{} (HCE) & 2115 & 2432 & 2476 & 2497 \\
\bottomrule
\end{tabular}
\vspace{-0.5em}
\caption{Performance of \ours{} with and without frontier-model distillation. The HCE variant uses an alternative data construction pipeline based on Stockfish's hand-crafted evaluation.}
\vspace{-1em}
\label{tab:hce}
\end{wraptable}

\paragraph{Moving away from frontier LM annotation.}
Finally, we investigate whether \ours{} can achieve strong performance without being seeded with explanations from a frontier model.
To this end, we construct an alternative training dataset from features derived from Stockfish's hand-crafted evaluation (HCE), using them to generate templatic explanations for each position.
We use this data to replace the \textsc{Sol}-seeded training round, while keeping the remainder of the training procedure unchanged. In \cref{tab:hce}, the HCE variant shows a similar trajectory of Elo improvement to \ours{}, suggesting that frontier-model distillation is not strictly necessary for \ours{} to learn to generate high-quality explanations. We report an example of its explanations and full details of the data-generation procedure in \cref{app:analysis}.

\section{Conclusion}

\ours{} generates higher-quality explanations than prior methods and frontier language models. Its move recommendations are more accurate, as reflected by substantially higher Elo ratings, and more strongly substantiated by the predicted principal variations. 
The prose of its explanations achieves coherence comparable to that of frontier language models, while remaining fluent.
Although we develop and evaluate \ours{} in the domain of chess, the underlying framework is not specific to it. 
Our architecture provides a general mechanism to couple a language model with a pretrained expert encoder, while our training procedure provides an iterative improvement algorithm for any stateful environment. 
Together, these components suggest a recipe for applying language models to domains where silent expert encoders are available, including games, robotics, and computer use. 
In such settings, expert models can provide the domain-specific representations and evaluations needed for strong decision-making, while language models can turn these signals into explanations that are accessible to humans and useful for further reasoning.

\subsection*{Acknowledgments}

We thank everyone in the Princeton Language Intelligence group for early discussions and feedback on drafts.
Jeff wishes to thank Yoonsang Lee and Simon Park for their many discussions regarding puzzles.
Adithya thanks Ijay Narang for chess games and feedback. Adithya and Jeff also look back fondly on the many chess games they played prior to and throughout the project as motivation.

\bibliography{iclr2027_conference}
\bibliographystyle{iclr2027_conference}

\newpage
\appendix

\section{Domain Adaptation Phase Details}
\label{app:domain}
This appendix section contains additional details regarding the domain adaptation stage. We discuss the data processing pipeline in \cref{app:domain_data}, the training process in \cref{app:domain_train}, and finally all architectural and decoder ablations in \cref{app:domain_ablations}.

\subsection{Data Processing}
\label{app:domain_data}

We first provide a list of all question types in each phase of the domain adaptation stage. 
As a reminder, the four phases are \textsc{Static-Current}, \textsc{Dynamic-Current}, \textsc{Static-Future}, and \textsc{Dynamic-Future}.
The first column gives the task while the second gives a sample response. All questions use our newly introduced tokens, and all answers also carry a machine parsable sequence used for checking correctness (omitted in this table).

{\small
\begin{longtable}{@{}p{4.6cm}p{9.4cm}@{}}
\toprule
\textbf{Question task} & \textbf{Sample Q\,$\rightarrow$\,A / answer clause} \\
\midrule
\endfirsthead

\toprule
\textbf{Question task} & \textbf{Sample Q\,$\rightarrow$\,A / answer clause} \\
\midrule
\endhead

\midrule
\multicolumn{2}{r@{}}{\textit{continued on next page}}\\
\endfoot

\bottomrule
\endlastfoot

\multicolumn{2}{@{}l}{\textbf{Stages 1 \& 3 --- Static Board Understanding}} \\
\midrule

Identify the piece on a queried square &
\texttt{What piece is on <SQUARE\_28>?} $\rightarrow$ \texttt{There is <PIECE\_MN> on <SQUARE\_28>.} \\

Locate all squares holding a queried piece &
\texttt{Where is <PIECE\_MR>?} $\rightarrow$ \texttt{<PIECE\_MR> occupies <SQUARE\_1> and <SQUARE\_8>.} \\

List pieces on a queried file &
\texttt{Pieces on file <SQUARE\_5>--<SQUARE\_61>?} $\rightarrow$ \texttt{1 <PIECE\_MP> and 1 <PIECE\_OK>.} \\

List pieces on a queried rank &
\texttt{Pieces on rank <SQUARE\_1>--<SQUARE\_8>?} $\rightarrow$ \texttt{Walking the rank: \ldots} \\

List pieces on a queried diagonal &
\texttt{Pieces on diagonal <SQUARE\_1>--<SQUARE\_64>?} $\rightarrow$ \texttt{\ldots (or ``open diagonal'')} \\

Count each piece type, both sides &
\texttt{Piece counts of both sides?} $\rightarrow$ \texttt{player: <PIECE\_MP>5<PIECE\_MN>2\ldots} \\

Total material points, both sides &
\texttt{Material points per side?} $\rightarrow$ \texttt{player 39, opponent 34.} \\

\midrule
\multicolumn{2}{@{}l}{\textbf{Stages 2 \& 4  --- Dynamic Move Transitions}} \\
\midrule

List legal moves of a queried piece &
\texttt{Legal moves for <PIECE\_MN> on <SQUARE\_28>?} $\rightarrow$ \texttt{<SQUARE\_43>, <SQUARE\_45>.} \\

List attackers/defenders of a square &
\texttt{Attack/defend <PIECE\_MP> on <SQUARE\_20>?} $\rightarrow$ \texttt{attacked by \ldots, defended by \ldots} \\

List every capture the mover can make &
\texttt{What captures can player make?} $\rightarrow$ \texttt{Player can capture with \ldots} \\

List every check the mover can give &
\texttt{What checks can player give?} $\rightarrow$ \texttt{The <PIECE\_OK> can be checked by \ldots} \\

List all legal ways out of check &
\texttt{How to get out of check?} $\rightarrow$ \texttt{capture / block / king-move \ldots} \\

Decide and explain checkmate &
\texttt{Is it checkmate?} $\rightarrow$ \texttt{Yes --- <PIECE\_MK> at <SQUARE\_2> is mated.} \\

\midrule

\caption{Question types used for our four-stage domain adaptation.}
\vspace{-0.5em}
\end{longtable}
}

We then provide the data composition for each phase of training. This data is provided in the following \cref{tab:domain_data}.

\begin{table}[H]
\centering
\begin{tabular}{l|cccc}
\toprule
Data source & \textsc{curr-static} & \textsc{curr-dynamic} & \textsc{future-static} & \textsc{future-dynamic} \\
\midrule
Stage-1 data & 100\% & -- & -- & --  \\
Stage-2 data & 10\% & 90\% & -- & -- \\
Stage-3 data & 6\% & 8\% & 86\% & -- \\
Stage-4 data & 6\%  & 6\% & 8\%  & 80\% \\
\bottomrule
\end{tabular}
\caption{Data composition of the four stages of domain adaptation training. Rows and columns correspond to the final stage data mixture and data sources described above. Earlier-stage data is replayed at a small proportion to prevent forgetting. }
\label{tab:domain_data}
\end{table}

\subsection{Training and Evaluation}
\label{app:domain_train}
The full training hyperparameters are shown below:

\begin{table}[H]
    \centering
    \begin{tabular}{lc}
        \toprule
        Hyperparameter & Value \\
        \midrule
        Training steps & \\
        \quad \textsc{stage 1} & 50K \\
        \quad \textsc{stage 2} & 60K \\
        \quad \textsc{stage 3} & 30K \\
        \quad \textsc{stage 4} & 100K \\
        Bridge learning rate & \\
        \quad \textsc{stage 1} & $1\times10^{-4}$ \\
        \quad \textsc{stage 2} & $8\times10^{-5}$ \\
        \quad \textsc{stage 3} & $7.5\times10^{-5}$ \\
        \quad \textsc{stage 4} & $2\times10^{-4}$ \\
        Token embedding learning rate & $0.1\times$ bridge LR \\
        Effective batch size & 256 \\
        Warmup & 5\% \\
        Final LR / peak LR & 5\% \\
        Precision & bf16 \\
        Optimizer & AdamW \\
        Learning-rate schedule & Cosine \\
        Early stopping & Yes \\
        \bottomrule
    \end{tabular}
    \caption{Training hyperparameters for the four-stage curriculum. Hyperparameters shared across all stages are listed once.}
    \label{tab:domain_hyperparameters}
\end{table}

And the results are here:

\begin{table}[H]
\centering
\small
\begin{tabular}{@{}cccc@{}}
\toprule
\multicolumn{2}{c}{\textsc{Static}} & \multicolumn{2}{c}{\textsc{Dynamic}} \\
\cmidrule(lr){1-2}\cmidrule(lr){3-4}
Current & Future & Current & Future \\
\midrule
0.9997 & 0.9997 & 0.9897 & 0.9607 \\
\bottomrule
\end{tabular}
\caption{Held-out test set accuracies of our model.}
\label{tab:domain_results}
\end{table}

\subsection{Ablations}
\label{app:domain_ablations}

We discuss three bridge architectures we tested, along with three decoder families below. Then, we provide the results in \cref{tab:domain_ablations}. As a reminder, our bridge serves to connect the BT5 network from Leela and a language model decoder.

\paragraph{Flamingo (gated cross-attention blocks).} Our Flamingo-inspired architecture inserts trainable gated cross-attention blocks between the usual decoder blocks \citep{flamingo}. We add 16 such blocks that pair the $i$th encoder hidden state with the $2i$th decoder hidden state, and are inserted before every other decoder layer (positions 0, 2, …, 30). We use all encoder hidden states so that multi-scale board features are progressively accumulated into the decoder's residual stream. The output of each gated cross-attention block is added to the residual stream through a $\tanh(\alpha)$ gate whose scalar $\alpha$ is initialized to zero, so cross-attention starts as a no-op and opens gradually, avoiding noise injection early in training. This bridge architecture contains about 470M parameters.

\paragraph{LLaVA (prefix token projection).} We tested a LLaVA-inspired architecture which instead injects the encoder hidden states by projecting them to the input embedding layer. We channel-concatenate all 16 encoder hidden states per square (64 × 16384), pass them through a two-layer MLP projector to the decoder's embedding dimension, and prepend the resulting 64 vectors as prefix "soft tokens" ahead of the text tokens. We add learned 2D file-and-rank embeddings to encode board geometry without disturbing the decoder's own positional embeddings. No new attention modules are introduced—the decoder's own self-attention attends over the injected prefix and text jointly. Whereas our Flamingo architecture froze the decoder parameters during domain adaptation, we found that we needed to unfreeze them to achieve meaningful training results; we adapt the decoder with LoRA (rank 16 on Q/K/V/O). This is by far the lightest bridge with only about 83M trainable parameters.

\textbf{KV Projection (Direct KV-cache Projection)} We try a third architecture that directly projects the concatenated encoder states into each decoder layer's key/value cache. A shared LayerNorm feeds per-layer $W_K$/$W_V$ maps that produce 64 key/value positions prepended to the \texttt{past\_key\_values} of every decoder layer, letting queries at each depth route toward encoder-derived context without any new attention sublayers or prefix tokens in the input sequence.  In initial testing this method performed poorly—it trailed both Flamingo and LLaVA by roughly 5–10 percentage points across every learning-rate and dataset configuration when trained on \textsc{Static} questions. It is also the heaviest bridge, containing approximately 612M trainable parameters. As a result, this bridge architecture was dropped from all subsequent experiments.

\textbf{Decoder Ablations} We consider three open pretrained LLMs of comparable scale: SmolLM3-3B, Qwen3-4B, and Gemma3-4B. We maintain the same training setup (data, optimizer, scheduler) across all configurations so ablations isolate the effects of the decoder language and the bridge architecture choices.
Family-specific tokenizer and stop-token handling (e.g. Gemma's \texttt{<end\_of\_turn>}) is managed by per-family adapters, keeping the split-embedding and training setup otherwise identical across backbones.

\begin{table}[H]
\centering
\small
\begin{tabular}{@{}llcccc@{}}
\toprule
& & \multicolumn{2}{c}{\textsc{Static}} & \multicolumn{2}{c}{\textsc{Dynamic}} \\
\cmidrule(lr){3-4}\cmidrule(lr){5-6}
\textbf{Architecture} & \textbf{Model family} & Current & Future & Current & Future \\
\midrule
\multirow{3}{*}{Flamingo}
  & SmolLM3 & 99.97 & 99.97 & 98.97 & 96.07 \\
  & Qwen3   & 99.80 & 99.87 & 98.42 & 96.83 \\
  & Gemma3  & 99.04 & 98.51 & 95.47 & 79.78 \\
\midrule
\multirow{3}{*}{LLaVA}
  & SmolLM3 & 99.80 & 99.80 & 99.00 & 93.20 \\
  & Qwen3   & 99.93 & 99.89 & 97.20 & 57.20$^{*}$ \\
  & Gemma3  & 99.87 & 99.23 & 95.17 & 50.70$^{*}$ \\
\bottomrule
\end{tabular}
\caption{Architecture $\times$ Model-family ablation: Accuracy (\%) on \textsc{Static} and \textsc{Dynamic} on held-out
validation sets.
$^{*}$Undertrained: the training run was aborted early, at step counts matched to Flamingo.}
\label{tab:domain_ablations}
\end{table}

Clearly, our Flamingo architecture outperforms LLaVA.  With the Flamingo bridge fixed, Gemma3 is clearly the weakest backbone—particularly on dynamic-future, where it drops to 79.8\%—leaving SmolLM3-3B and Qwen3-4B as the real contenders. SmolLM3 is the more efficient choice on two counts: it is smaller and is more accurate on three of the four tasks that we evaluated. Thus, we adopt SmolLM3-3B as our decoder.

\section{Iterative Improvement Phase Details}
\label{app:iterative}

\subsection{Data Processing}
\label{subapp:datacreation}

We provide here all the little details that go into the data-creation process of each stage of iterative self-distillation.
Below, ``model'' refers to the latest checkpoint \intermediate-$k$, which is used as described to create SFT data for \intermediate-$(k+1)$  (hyperparameters in Appendix~\ref{subapp:iterativeparams}).

\paragraph{The pool of seed positions.}
Our pool of seed positions comes from three sources, and we always deduplicate against our two benchmark positions to avoid contamination:
\begin{enumerate}
    \item We select 210K positions from human play. The games are disjoint from those used in previous iterations, but positions may recur. We select at most 6 positions from a given game.
    \item 115K positions come from the Lichess puzzles, where we ensure disjointness with those used in previous iterations.
    \item We run 6250 chess games between the model and Stockfish, with colors chosen at random, and the number of nodes used by Stockfish drawn from log-random(100, 100K). We find that skewed positions (Stockfish evaluation over +2.5 or under -2.5) are over-represented, and we downsample them by a factor of $5\times$. Then, all positions where the side-to-play made a mistake per Stockfish (100K nodes) are retained, along with a random sample of 12.5\% of the rest. This yields between 50-100K positions.
\end{enumerate}

Using these positions as the pool of ``seed'' positions, we apply Algorithm~\ref{alg:iterative_improve} to mine data for training. In particular, we do:

\paragraph{Recursive sampling.}
Given a seed position -- which we refer to as the \emph{root} -- we run the model analysis to obtain three promising moves. If all three moves are mistakes as per Stockfish (as per a 5\% winrate cutoff), we replace the worst one with the best Stockfish move.
We apply the three moves to the position to obtain three \emph{children}, which are themselves analyzed by our model.
Now, if any of the three analyses recommend a mistake (or illegal move), or their evaluation diverges from that of Stockfish at the child by more than 10\% win-rate, we redo the entire process with the offending child as the new root.
The rationale here is that we recursively descend into model mistakes until the root is beyond the capability of the model but the children are not.
We give up recursing after five attempts.

\paragraph{Consolidation.}
Once we have found the true root as per the process above, we pass the three child explanations to Qwen3.8-27B, which is chosen for its ability to strictly adhere to provided instructions and compose the given explanations into a root explanation grounded entirely in them.
We use the prompt from Appendix~\ref{app:prompts} towards this end.
Finally, we overwrite just the numerical component of the evaluation section with the true Stockfish evaluation (100K nodes) which provides better search calibration for future iterations.

\paragraph{Training.}
We collect all roots and consolidated explanations into a dataset, and discard those where the supplied PV is worse than the PV of the original model applied to the root. 
This is determined by comparing the two moves at the first divergence of the two PVs. 
The final yield of the entire process is usually around 250K examples, though the data generated for \intermediate{}-2 only had 150K instances, as the initial seeded model often made mistakes.
Finally, the next iteration of the model is trained using SFT on the resulting dataset, with the hyperparameters in Appendix~\ref{subapp:iterativeparams}.

\begin{algorithm}[t]
\caption{Data creation for iterative self-distillation}
\label{alg:iterative_improve}
\begin{algorithmic}[1] %

\Procedure{ImprovesPrincipalVariation}{Root, PVBefore, PVAfter}
    \State Position, MoveBefore, MoveAfter $\gets$ FindFirstDivergence(Root, PVBefore, PVAfter)
    \If{MoveAfter is better than MoveBefore} \Comment{Compare using Stockfish}
    \State \textbf{return} True
    \Else
    \State \textbf{return} False
    \EndIf
\EndProcedure\\

\Procedure{MinePosition}{ChessLM, Root} \Comment{Find a position to train on}
    \State RootEval, CandidateMoves, PV, Analysis $\gets$ ChessLM(Root) 
    \State CandidateMoves $\gets$ SortByQuality(CandidateMoves)
    \If{all moves are inaccuracies or worse} \Comment{5\% drop in winrate per Stockfish}
        \State CandidateMoves $\gets$ Stockfish(Root) $\cup$ CandidateMoves[:-1] \Comment{Inject best move}
    \EndIf\\
    \State Children $\gets$ Apply(Root, CandidateMoves)
    \State ChildrenEval, ChildrenCandidates, ChildrenPV, ChildrenAnalysis $\gets$ ChessLM(Children)
    \State ChosenMoves $\gets$ ChildrenCandidates[:, 0] \Comment{Find model's preferred move}\\
    
    \If{any of ChildrenEval or ChosenMoves is a mistake} \Comment{10\% drop in winrate}
        \State OffendingChild $\gets$ GetArbitraryOffendingChild(Children)
        \State \textbf{return} \Call{MinePosition}{ChessLM, OffendingChild} \Comment{Recurse into mistakes}
    \EndIf\\

    \State TopChild, TopMove, TopPV $\gets$ ChooseTop(Children) \Comment{Select move w/ highest ChessLM eval}
    \State InferredPV $\gets$ TopMove $\cup$ TopPV \Comment{Construct principal variation}
    \If{\Call{ImprovesPrincipalVariation}{Root, PV, InferredPV}} 
    \State \textbf{return} Root, Consolidate(ChildrenAnalysis) \Comment{Use Qwen3.8-27B to combine analyses}
    \Else
    \State \textbf{return} NULL, NULL \Comment{Drop positions where search doesn't improve over the root}
    \EndIf
\EndProcedure\\

\Procedure{GetTrainingData}{ChessLM, PositionPool}
\State Data $\gets \emptyset$
\ForAll{Root $\in$ PositionPool}
    \State NewRoot, NewAnalysis $\gets$ \Call{MinePosition}{ChessLM, Root}
    \If{NewAnalysis is not NULL}
    \State Data $\gets$ Data $\cup$ \{(NewRoot, NewAnalysis)\}
    \EndIf
\EndFor
\State \textbf{return} Data
\EndProcedure

\end{algorithmic}
\end{algorithm}

\newpage
\subsection{Hyperparameters for iterative improvement}
\label{subapp:iterativeparams}
In this section, we provide the hyperparameters used for SFT for (a) the initial distillation step we use to seed the model with analyses from GPT-5.6-Sol (low), and (b) the SFT used for each iteration of the self-distillation process.

\paragraph{Hyperparameters for the initial seeding (distillation) from GPT-5.6-Sol (low).}
We use 11,250 positions from general human play and 3,750 puzzle positions as our pool of positions, for which we prompt GPT-5.6-Sol (low) with the prompt shown in Appendix~\ref{app:prompts} -- costing us \$652.35.
We then re-prompt GPT-5.6-Sol (low) (prompt in Appendix~\ref{app:prompts}) to rewrite its response in special tokens of the form \texttt{<WHITE\_PAWN>}. Note that these are not our special POV-tokens, but can be replaced one-to-one with them, which allows us to not have GPT juggle the reflection of the board.
We select only those positions where the supplied analysis recommends a move that is legal and drops at most 10\% win rate as per Stockfish at 100,000 nodes -- bringing the count to 8602 positions. We train on these positions for 4 epochs, as the hyperparameters in Table~\ref{tab:sol_sft_hyperparameters} show.

\begin{table}[H]
  \centering
  \begin{tabular}{lc}
      \toprule
      Hyperparameter & Value \\
      \midrule
      Training examples & 8,402 \\
      Validation examples & 200 \\
      Training steps & 2,100 \\
      Effective batch size & 16 \\
      Maximum sequence length & 2,048 \\
      Bridge learning rate & $1\times10^{-5}$ \\
      Decoder learning rate & $1\times10^{-6}$ \\
      Token embedding learning rate & $1\times10^{-6}$ \\
      Warmup & 5\% \\
      Final LR / peak LR & 100\% \\
      Precision & bf16 \\
      Optimizer & AdamW \\
      Weight decay & 0.1 \\
      Gradient clipping norm & 1.0 \\
      \bottomrule
  \end{tabular}
  \caption{Supervised fine-tuning hyperparameters for distillation
  from GPT-5.6-Sol (low).}
  \label{tab:sol_sft_hyperparameters}
\end{table}

\paragraph{Hyperparameters for the SFT used for each self-distillation iteration.}
As explained in Appendix~\ref{subapp:datacreation}, the final number of examples we use for training in each iteration ranges from 150K (first iteration, when most instances are rejected) to 283K. 
Regardless, we standardize other hyperparameters as in Table~\ref{tab:self_improvement_sft_hyperparameters}.

\begin{table}[H]
  \centering
  \begin{tabular}{lc}
      \toprule
      Hyperparameter & Value \\
      \midrule
      Training examples & 150K--285K \\
      Maximum training steps & 30K \\
      Effective batch size & 32 \\
      Maximum sequence length & 2,048 \\
      Bridge learning rate & $1\times10^{-4}$ \\
      Decoder learning rate & $1\times10^{-5}$ \\
      Token embedding learning rate & $1\times10^{-5}$ \\
      Warmup & 5\% \\
      Precision & bf16 \\
      Optimizer & AdamW \\
      Weight decay & 0.1 \\
      Gradient clipping norm & 1.0 \\
      Learning-rate schedule & Constant after warmup \\
      \bottomrule
  \end{tabular}
  \caption{Supervised fine-tuning hyperparameters for one
  self-distillation iteration.
  Each iteration initializes from the selected checkpoint of
  the preceding iteration.}
  \label{tab:self_improvement_sft_hyperparameters}
\end{table}

\section{Evaluation Details}
\label{app:evals}

\paragraph{Accuracy Evaluations on Full Games.} 
Each evaluated model plays 4 games against each of the eight opponents listed in Table~\ref{tab:elo-opponents}. The four games include two each (one with each color) after the opening books \textbf{1. e4 e5 2. Nf3 Nc6} and \textbf{1. d4 d5 2. c4 e6}. The former leads to openings such as the Ruy Lopez and the Italian, while the latter is a Queen's Gambit Declined. 
We would like to run even more games, but running games with frontier LMs is quite expensive, often costing north of \$15 per game with GPT-5.6-Sol (high).

\begin{table}[t]
  \centering
  \small
  \begin{tabular}{lrr}
      \toprule
      \textbf{Opponent} & \textbf{Search budget (nodes)} & \textbf{Anchored Elo}
      \\
      \midrule
      Stockfish                  & 10      & 1978 \\
      Maia-2200 (Lc0)             & 1       & 2037 \\
      Stockfish                  & 100     & 2085 \\
      TinyGyal-8 (Lc0)            & 10,000  & 2422 \\
      Stockfish                  & 1,000   & 2597 \\
      11258-16x2-se-4 (Lc0)       & 10,000  & 2689 \\
      Stockfish                  & 3,000   & 2838 \\
      11258-24x3-se (Lc0)         & 10,000  & 2865 \\
      \bottomrule
  \end{tabular}
  \caption{The opponents we use for accuracy evaluations in Section~\ref{sec:accuracy}.}
  \label{tab:elo-opponents}
\end{table}

\paragraph{Substantiation Evaluations on Held Out Positions.} 
All our evaluations are performed with Stockfish with 1M nodes of search.

\paragraph{Coherence Evaluations with LM Annotations.}
We perform the LM Judge annotation with GPT-5.6-Sol (high) with the prompt in \cref{app:prompts}.
All Stockfish annotations passed to the judge are performed at 1M nodes.

\section{Analysis Details}
\label{app:analysis}

\paragraph{Moving away from frontier LM annotation.}
We construct the seed explanations for \ours{} (HCE) by combining engine search with templated descriptions of handcrafted evaluation features. Each explanation considers several candidate moves, follows their continuations, and explains why some lines are inferior before identifying a best move and critical line. No frontier LM is used to write these seed explanations. We describe the construction below, distinguishing the numerical evaluator used for search from the handcrafted features used to verbalize it. The examples are provided in \cref{app:examples}.

\paragraph{Constructing the analysis tree.}
Starting from a root position, we run alpha--beta search to an initial depth of three ply. We use Stockfish's neural network to score positions without further search. Captures are searched before other moves, prioritizing valuable captured pieces and inexpensive attackers. 
After reaching depth three, we continue searching captures and checks so that a line does not stop immediately before a forcing exchange. We consider non-capturing checks for the first two additional plies and captures for up to eight additional plies. When the side to move is not in check, we also allow the current evaluation to stand without extending the line. When it is in check, we instead search legal replies that escape check. Checkmate receives a mate score, while stalemate, insufficient material, and claimable draws receive zero. Thus, three ply specifies the initial search depth, not a fixed length for the final explanation.

\paragraph{Including plausible mistakes.}
An explanation containing only best play would provide little supervision for rejecting tempting alternatives. We therefore use the Maia-1100 branch of Lc0 networks to introduce plausible human mistakes into the analysis. Alongside the search-preferred moves, we include alternatives suggested by Maia and their responses, allowing the explanation to show why a tempting move is inferior.
For the HCE seed dataset, we target a retained tree of 40 nodes. This controls initial expansion rather than imposing a hard limit on the final tree: the judging pass below may extend or repair its lines.

\paragraph{Judging and completing variations.}
Evaluations without search are useful for constructing the tree but can miss tactical consequences. We therefore use a separate Stockfish search with a budget of 1,000 nodes per queried position to judge the retained continuations. We clear the engine's search memory between new queries to avoid dependence on query order. We work backward from the final positions, with each player choosing the continuation that is best from its own perspective.

A loss of at least 10 percentage points in expected win rate relative to the best available continuation is considered a mistake. If all retained moves at a node meet this threshold relative to the judge's evaluation, and its preferred move is missing, we insert the judge's PV. We extend this inserted line to the depth of the shortest existing continuation when the PV permits. If the existing subtree is only a single continuation, we replace that continuation rather than retaining a forced mistake. Finally, if the judge's preferred move at the end of a line is a capture or check, we extend the line by up to two PV plies and judge again, for at most six such extensions. This reduces explanations that end in the middle of a tactical sequence.

\paragraph{Refutations and supersession.}
We narrate one branch at a time, considering checks before captures, threats, and other moves. As variations are completed, we compare their outcomes from the perspective of the player choosing where the lines diverge. A line containing a move that loses at least 10 percentage points under the resulting evaluations is marked as suboptimal or refuted. Other lines receive numbered labels and can be \emph{superseded} by a preferable alternative.

A supersession is stated only when the explored branches justify it. In particular, all retained opponent alternatives along the winning branch must have been presented, as must all retained alternatives for the choosing player along the losing branch. Otherwise, a later reply could invalidate the comparison. We resolve comparisons at deeper branch points before shallower ones, and break equal-value ties by presentation order. A line that has itself been superseded may still support a comparison strictly below the branch point where it lost; it is explicitly identified as rejected when used this way. The concluding critical line follows the surviving best-play continuation. These comparisons establish preference within the constructed tree, rather than claiming an exhaustive proof over all legal continuations.

\paragraph{Verbalizing the analysis.}
For every narrated move, we compute the change in handcrafted features, including material, pawn structure, piece activity, mobility, king safety, threats, passed pawns, space, and initiative. We rank feature changes by their absolute evaluation contribution and verbalize up to three salient effects using templates. Additional tactical pattern checks annotate clearly preferred moves when the Stockfish judge estimates a best-versus-second-best gap of at least five win-rate percentage points. Connective templates indicate when the analysis returns to an earlier position, considers another reply, refutes a move, or supersedes a previously discussed line. Repeated refutations are compressed by describing their shared consequence once and then highlighting the differences. The explanation ends with the critical variation and its recommended first move, expressed in the same player-relative chess vocabulary used elsewhere in training.
Below is a short example from the HCE seed dataset, with chess tokens decoded into readable piece and square names. This example follows a single retained continuation; longer examples additionally compare alternatives using the refutation and supersession statements described above.

\begin{tcolorbox}[
  title={Example HCE seed analysis (Black to move)},
  breakable,
  colback=white,
  colframe=black!60,
  fontupper=\small
]
It is our move.

\medskip
Consider black knight on g8 advances to f6. It centralizes the black knight, leaves the opponent's white pawn on e4 hanging (undefended), and gives the black knight now on f6 more scope. The opponent replies white pawn on d2 moves to d3, which defends the opponent's white pawn on e4 and connects the white pawn on c2 and d3 and connects the white pawn on d3 and e4. We continue with black queen on d8 advances to e7. I'll note this position as [1] for later.

\medskip
Let me recall the best variation I have found, which leads to position [1]: it runs 1... black knight g8-f6 2. white pawn d2-d3 black queen d8-e7. Therefore, I should play \textbf{black knight on g8 advances to f6}.
\end{tcolorbox}

\section{Example Outputs}
\label{app:examples}

\begin{center}
\includegraphics[width=0.36\linewidth]{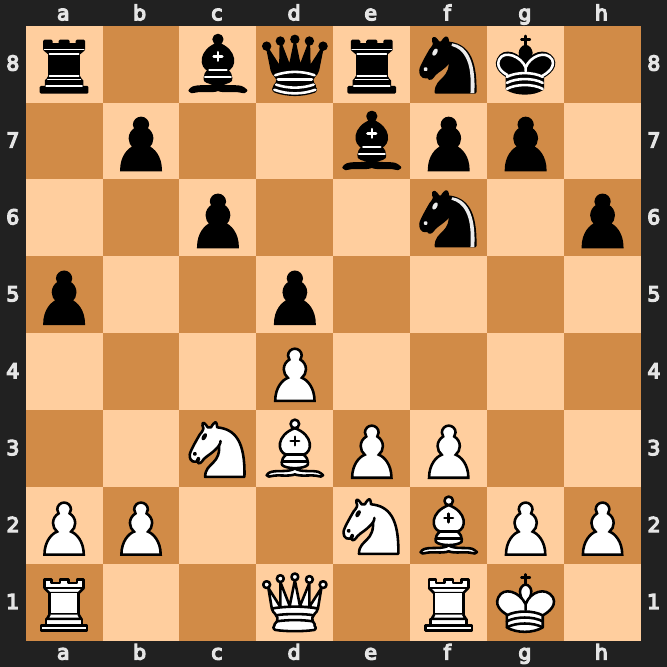}

\small White to move.
\end{center}

\subsection{\ours{}}

\begingroup
\renewcommand{\ttdefault}{pcr}
\begin{tcolorbox}[title={\ours{}},breakable,colback=white,colframe=black!60,fontupper=\small]
\noindent \texttt{ANALYSIS:}\\
\texttt{The root position features a tense central structure where the player holds a pawn duo on \textbf{d4} and \textbf{e3}, supported by the white bishop on \textbf{d3} and the white knight on \textbf{c3}. The opponent possesses the bishop pair, with the black bishop on \textbf{e7} and the black knight on \textbf{f6} well-placed, but the black bishop on \textbf{c8} remains undeveloped and the black knight on \textbf{f8} is still on its starting square. The key strategic question is how the player can consolidate the central space advantage and complete development before the opponent activates the light-squared bishop.}\par
\medskip

\noindent \texttt{The best move is \textbf{13. Qc2}, centralizing the queen on a useful diagonal where it supports the center and prepares to connect the rooks. The opponent's most accurate reply is \textbf{13... Bd6}, activating the bishop to a strong central square where it eyes \textbf{h2} and supports the center. The player's most effective continuation is \textbf{14. Rad1}, connecting the rooks and preparing to double on the \textbf{d-file} or redirect to the center. The opponent should then play \textbf{14... Be6}, developing the last minor piece to an active diagonal where it eyes \textbf{c4} and \textbf{d5}. The player's most purposeful follow-up is \textbf{15. e4}, gaining central space and creating a powerful pawn duo on \textbf{d4} and \textbf{e4}. The opponent's most accurate response is \textbf{15... dxe4}, resolving the central tension by eliminating the advanced pawn and opening the \textbf{e-file}. After \textbf{16. fxe4}, the player retains a modest space advantage with the pawn on \textbf{e4}, while the opponent has active pieces and can continue development with \textbf{...N8d7}, preparing \textbf{...Nb6} to challenge the white bishop on \textbf{d3}. The position is roughly balanced with a small edge to the player.}\par
\medskip

\noindent \texttt{The alternative \textbf{13. e4} is inferior. The opponent responds with \textbf{13... dxe4}, resolving the central tension in the opponent's favor. After \textbf{14. fxe4} \textbf{14... Ng4}, the knight reaches a strong outpost where it attacks the white bishop on \textbf{f2} and creates immediate tactical threats. The player's most accurate response is \textbf{15. Qd2}, but the opponent continues with \textbf{15... Nxf2}, capturing the bishop and forcing a recapture. After \textbf{16. Rxf2} \textbf{16... Bd6}, the opponent develops with tempo against the white rook and prepares \textbf{...Ne6}. The opponent retains the bishop pair and a modest positional edge of approximately 0.32 pawns.}\par
\medskip

\noindent \texttt{The candidate \textbf{13. a3} is a quiet flank move that does not address the central tension. The opponent responds with \textbf{13... Bd6}, and after \textbf{14. Qc2} \textbf{14... Ne6} \textbf{15. Rad1} \textbf{15... Bd7} \textbf{16. Rfe1}, the position is approximately equal with the player retaining a small space advantage but the opponent having adequate counterplay and the bishop pair.}\par
\medskip

\noindent \texttt{BEST\_MOVE: \textbf{Qc2}}\\
\texttt{CRITICAL\_LINE: \textbf{Qc2} \textbf{Bd6} \textbf{Rad1} \textbf{Be6} \textbf{e4} \textbf{dxe4} \textbf{fxe4}}\\
\texttt{PROMISING\_MOVES: \textbf{Qc2}, \textbf{a3}, \textbf{e4}}\\
\texttt{EVALUATION: the player holds a modest advantage of approximately 0.54 pawns, with a central space edge and active piece coordination after best play.}\par
\end{tcolorbox}

\subsection{\ours(HCE)}

\begin{tcolorbox}[title={\ours{} \textsc{(HCE)}},breakable,colback=white,colframe=black!60,fontupper=\small]
\noindent \texttt{ANALYSIS:}\\
\texttt{White holds a modest spatial advantage in a closed-center structure. The white pawn on \textbf{d4} and \textbf{e3} form a compact chain supported by the white knight on \textbf{c3} and white bishop on \textbf{d3}, while Black's black pawn on \textbf{d5} and black knight on \textbf{f6} provide the counterplay framework. The key dynamic tension lies in the center: White's natural plan is \textbf{e4} to open lines and activate the white bishop on \textbf{f2}, while Black's counterplay revolves around \textbf{...c5}, \textbf{...b5}, and the \textbf{...Ng4} knight sortie. The practical question is how White should prepare for the central break while maintaining flexibility.}\par
\medskip

\noindent \texttt{I. \textbf{Qc2} — the best move. The queen supports the \textbf{e4} break, keeps the white bishop on \textbf{f2} flexible, and avoids committing the rook to the \textbf{d-file} prematurely. Black's best reply is \textbf{1... Bd6}, placing the bishop on the long diagonal where it eyes \textbf{h2} and supports the \textbf{...c5} break. White's most direct plan is \textbf{2. e4}, challenging the center. After \textbf{2... dxe4} \textbf{3. fxe4}, the white pawn on \textbf{e4} is a strong central presence, but Black's black bishop on \textbf{d6} is well placed and the position remains balanced with a slight White edge. Black's alternative \textbf{2... Ng4} is inferior: \textbf{3. e5} attacks the black bishop on \textbf{d6} and gains space, and after \textbf{3... Be7} \textbf{4. h3}, the knight is kicked with no compensation. If White instead plays \textbf{2. h3} prophylactically, Black responds \textbf{2... b5}, freeing the black bishop on \textbf{c8} and preparing \textbf{...b4} to dislodge the white knight on \textbf{c3}; after \textbf{3. e4} \textbf{dxe4} \textbf{4. fxe4}, the position is roughly equal. The slow \textbf{2. a3} allows \textbf{2... Ng4} with pressure on \textbf{e3}, and after \textbf{3. e4} \textbf{dxe4} \textbf{4. fxe4}, the black knight on \textbf{g4} is well placed and the position favors Black slightly.}\par
\medskip

\noindent \texttt{II. \textbf{Rc1} — a solid developing move that connects the rooks and prepares \textbf{e4}, but it is slightly less flexible than I because the rook on \textbf{c1} is less useful than the queen on \textbf{c2} in supporting the central break. Black's best reply is \textbf{1... b5}, immediately challenging the white knight on \textbf{c3} and preparing \textbf{...b4}. White's strongest response is \textbf{2. e4}, opening the center. After \textbf{2... dxe4} \textbf{3. fxe4}, White retains a comfortable spatial advantage with the white pawn on \textbf{e4} and the open \textbf{f-file}, but Black's queenside counterplay provides sufficient compensation. The alternative \textbf{2. a3} is prophylactic but concedes the initiative; after \textbf{2... b4} \textbf{3. Na4}, the knight is displaced and Black's queenside space is more effective. The passive \textbf{2. Qc2} allows \textbf{2... b4} \textbf{3. Na4} \textbf{c5}, connecting the queenside pawns and preparing \textbf{...c4} to dislodge the white bishop on \textbf{d3}.}\par
\medskip

\noindent \texttt{III. \textbf{e4} — the immediate central break is premature and allows Black to equalize. Black's best reply is \textbf{1... dxe4}, resolving the tension on Black's terms and opening the \textbf{e-file}. After \textbf{2. fxe4} \textbf{Ng4} \textbf{3. Bg3} \textbf{Bd6}, the position is approximately equal: White retains the bishop pair and a solid center, while Black has the open \textbf{e-file} and active piece play. The alternative \textbf{2. Bxe4} is inferior: after \textbf{2... Nxe4} \textbf{3. fxe4} \textbf{Bd6}, Black obtains the bishop pair and a comfortable game. The retreat \textbf{2. Bc2} is a serious error: \textbf{2... exf3} wins a pawn and opens the \textbf{f-file}, and after \textbf{3. Rxf3} \textbf{Ng4}, Black has a clear advantage with the extra pawn and active pieces.}\par
\medskip

\noindent \texttt{BEST\_MOVE: \textbf{Qc2}}\\
\texttt{CRITICAL\_LINE: \textbf{Qc2} \textbf{Bd6} \textbf{e4} \textbf{dxe4} \textbf{fxe4}}\\
\texttt{PROMISING\_MOVES: \textbf{Qc2}, \textbf{Rc1}, \textbf{e4}}\\
\texttt{EVALUATION: the player has a modest advantage of 0.5 pawns.}\par
\end{tcolorbox}
\endgroup

\section{Prompts}
\label{app:prompts}

We list below the prompts we use for various purposes. First, the prompt used with GPT-5.6-Sol (low) to generate seed analyses, and also for the evaluation of various GPT and Gemini models across our benchmarks:

\begin{tcolorbox}[
  title={Prompt for seed-distillation},
  colback=white,
  colframe=black!60,
  fontupper=\small
]
\textbf{Instructions}

Analyze the supplied chess position in detail. Discuss the important features
of the position, candidate moves, and the concrete variations needed to decide
on the best move. Do not assume that the position is a tactical puzzle.

\medskip
Identify the best move for the side to move and give one principal critical
line assuming best play. Then identify the legal root moves that look most
promising and would deserve further investigation in a deeper search. Give at
least one such move, normally two or three, and never more than four; include
the best move among them and order them from most to least promising.

\medskip
Finally, evaluate the current position in a sentence containing a numerical
estimate. Use pawn units from White's perspective, where positive favors White,
negative favors Black, and 0.00 is equal. If there is a forced mate, a signed
mate distance may replace the pawn estimate.

\medskip
Use standard UCI coordinate notation for every structured move. Every root
move must be legal in the supplied position. The critical line must begin with
the stated best move and contain only moves that are legal in sequence.

\medskip
Return exactly:

\begin{flushleft}
\ttfamily\footnotesize
ANALYSIS:\\
<your detailed analysis>\\
BEST\_MOVE\_UCI: <move>\\
CRITICAL\_LINE\_UCI: <move1> <move2> <move3> ...\\
PROMISING\_MOVES\_UCI: <move1> [<move2> <move3> <move4>]\\
EVALUATION: <one sentence containing a numerical estimate>
\end{flushleft}

There must be exactly one of each labeled field and no text after EVALUATION.

\tcblower
\textbf{Per-position input}

\medskip
Previous position FENs, oldest to newest
(the current position is not included):\\
\texttt{<numbered history FENs, or (none available)>}

\medskip
Current FEN: \texttt{<FEN>}\\
Side to move: \texttt{<White or Black>}
\end{tcolorbox}

and the prompt used to translate these analyses into our special tokens:

\begin{tcolorbox}[
  title={Translation prompt for seed-distillation data},
  breakable,
  colback=white,
  colframe=black!60,
  fontupper=\small
]
You are a lossless translator from ordinary chess prose into a fixed
machine-readable chess vocabulary.

\medskip
The input contains a FEN followed by a five-field chess analysis.
Reproduce the five-field analysis exactly, changing nothing except
chess references in the prose under ANALYSIS and EVALUATION.
Copy BEST\_MOVE\_UCI, CRITICAL\_LINE\_UCI, and PROMISING\_MOVES\_UCI
exactly byte for byte. Do not tokenize, rewrite, reorder, correct,
or reinterpret any UCI move.

\medskip
\textbf{ALLOWED TOKENS}
\begin{itemize}
  \item Pieces: \texttt{<WHITE\_PAWN>},
  \texttt{<WHITE\_KNIGHT>}, \texttt{<WHITE\_BISHOP>},
  \texttt{<WHITE\_ROOK>}, \texttt{<WHITE\_QUEEN>},
  \texttt{<WHITE\_KING>}, and the corresponding six
  \texttt{<BLACK\_...>} tokens. Every mention of a specific or
  generic chessman uses the correctly colored token, including
  plurals and phrases such as ``bishop pair''. Infer the color
  from the position and local variation.

  \item Squares: \texttt{<SQUARE\_E4>}, with an uppercase file
  and rank.

  \item Files: \texttt{<FILE\_E1\_E8>}, using the two endpoint
  squares of the file.

  \item Ranks: \texttt{<RANK\_A4\_H4>}, using the two endpoint
  squares of the rank.

  \item Diagonals: \texttt{<DIAGONAL\_A1\_H8>}, using the two
  endpoint squares of the complete diagonal and putting the
  alphabetically earlier endpoint first.

  \item Players/colors: \texttt{<WHITE>} and \texttt{<BLACK>}.
  Replace every reference to either player or color as such,
  including white, black, White's, Black's, the side to move,
  the opponent, they, their, we, our, and similar references,
  with the correct absolute-color token. A color attached to
  a piece is represented only by its piece token:
  ``White's queen'' becomes \texttt{<WHITE\_QUEEN>}, not
  \texttt{<WHITE>}'s \texttt{<WHITE\_QUEEN>}.
\end{itemize}

\textbf{MOVES IN PROSE}
\begin{itemize}
  \item A non-capture is
  \texttt{<PIECE> <FROM\_SQUARE> <TO\_SQUARE>}.
  \item A capture is
  \texttt{<PIECE> <FROM\_SQUARE> <CAPTURED\_PIECE> <TO\_SQUARE>}.
  \item Castling is the king move. A promotion appends the
  promoted piece token.
  \item Resolve abbreviated SAN and natural-language moves
  using the FEN and the variation context.
\end{itemize}

\textbf{MOVE NUMBERS}

Move numbers in the supplied analysis have already been normalized
relative to the current FEN: the first move is numbered 1. when
White is to move or 1... when Black is to move. Preserve every
supplied move number and ellipsis exactly. Do not restore game
move numbers.

\medskip
SUBSTITUTE IN PLACE. Preserve every non-chess word, all punctuation,
paragraph breaks, labels, normalized move numbers, and numerical
evaluations. Do not summarize, improve, correct, or add commentary.
Do not add Markdown fences. Never invent a token. The tokens
listed above are the only text that may appear in angle brackets;
no other extraneous tokens are permitted. Output only the translated
five-field analysis.

\tcblower
\textbf{Per-position input}

\medskip
FEN: \texttt{<FEN>}

\medskip
Original five-field analysis:\\
\texttt{<analysis with normalized move numbers>}
\end{tcolorbox}

When training and evaluating \ours{}, we use the following shorter prompt to the LM decoder:
\begin{tcolorbox}[
  title={Training prompt for \ours{}},
  breakable,
  colback=white,
  colframe=black!60,
  fontupper=\small
]
Analyze the given chess position, identify the best move, give the
critical line assuming best play, list promising moves for deeper
investigation, and evaluate the position.
\end{tcolorbox}
For consolidation, Qwen3.8-27B is given the following prompt:
\begin{tcolorbox}[
  title={Qwen3.8-27B consolidation prompt},
  breakable,
  colback=white,
  colframe=black!60,
  fontupper=\small
]
You are consolidating three recursive evaluations of a
single root chess position. You are a faithful editor and synthesizer,
not an independent chess analyst. Every conclusion, positional claim,
candidate, tactic, and move sequence in your answer must be grounded
in the three supplied child analyses. Do not add chess knowledge or
repair their analysis yourself.

\medskip
Each child position was reached by playing the labeled candidate
from the root. The analysis model then began afresh: its move numbers
restart at 1 or 1..., and its side to move is the root player's
opponent. Every child evaluation is therefore from the OPPONENT'S
perspective, not the root player's perspective.

\medskip
The numerical selection rule supplied in the input is mandatory.
Choose the root candidate with the LOWEST signed numerical evaluation
for the opponent. For example, -2.25 for the opponent is better for
the root player than +1.03 or +2.68 for the opponent. Equivalently,
negate the child evaluation to obtain the root player's evaluation,
then choose the highest result. Treat -M as lower than every pawn
value and +M as higher than every pawn value. Do not override this
arithmetic rule using the prose, even if a child's narrative sounds
more favorable. BEST\_MOVE must be exactly the candidate marked
MANDATORY BEST ROOT CANDIDATE in the input.

\medskip
In the consolidated prose and structured fields, restore the root
player's point of view and restart move numbering from the root.
A continuation through a child must begin with the labeled root
candidate before the child's line.

\medskip
Preserve the established explanatory style and order. Begin with
general considerations and important features of the root position
and its candidates. Use later paragraphs for increasingly concrete
plans, comparisons, tactics, logic, and variations. Condense repeated
motifs such as a common fork, outpost, or maneuver, but preserve
decisive tactics and forcing lines at full useful length, even when
they are long. Emphasize the best candidate more than inferior ones.
The result is a root analysis, not a broad summary of three positions.

\medskip
Use only this absolute machine vocabulary:
\begin{itemize}
  \item squares: \texttt{<SQUARE\_E4>};
  \item pieces: \texttt{<WHITE\_PAWN>},
  \texttt{<BLACK\_BISHOP>}, and analogous colored piece tokens;
  \item players: \texttt{<WHITE>} and \texttt{<BLACK>};
  \item files, ranks, and diagonals: \texttt{<FILE\_E>},
  \texttt{<RANK\_4>}, \texttt{<DIAGONAL\_A1\_H8>}.
\end{itemize}
Never emit POV tokens such as \texttt{<PIECE\_MN>} or numbered
squares such as \texttt{<SQUARE\_29>}.

\medskip
Every complete move is written as adjacent tokens. A non-capture
has exactly three tokens,
\texttt{<PIECE><FROM\_SQUARE><TO\_SQUARE>}.
A capture has exactly four,
\texttt{<PIECE><FROM\_SQUARE><CAPTURED\_PIECE><TO\_SQUARE>}.
A promotion appends the promoted piece as a rare fifth token.
Apply this convention both in prose and in the structured fields.

\medskip
Return exactly these five fields and no surrounding commentary:
\begin{flushleft}
\ttfamily\footnotesize
ANALYSIS:\\
<fluent grounded root analysis>\\
BEST\_MOVE: <one tokenized root move>\\
CRITICAL\_LINE: <one tokenized root line>\\
PROMISING\_MOVES: <one to four tokenized legal root candidates, best first>\\
EVALUATION: <one sentence with a pawn or mate evaluation from the root player's
perspective>
\end{flushleft}

You have the complete three child analyses. Consolidate them to
approximately the length of one individual child analysis: compress
overlap and inferior branches to make room for the most important
concrete detail.

\medskip
Every CHILD CRITICAL\_LINE below has been mechanically truncated
immediately before its first illegal move or its first move with a
Stockfish-100k win-rate drop of at least 10 percentage points.
Each child block states the exact boundary. When a line was
truncated, omit the rejected move and every continuation that
depends on it from both the consolidated prose and structured
CRITICAL\_LINE. Do not repair, reconstruct, extend, or independently
analyze that rejected continuation. You may retain and use the
supplied clean prefix.

\medskip
The consolidated CRITICAL\_LINE must begin with the MANDATORY BEST
ROOT CANDIDATE and may continue only with a prefix of that candidate's
supplied clean child CRITICAL\_LINE. If no child continuation remains,
output only the root candidate. PROMISING\_MOVES may contain only
the supplied legal root candidates. The original root analysis is
deliberately not supplied: combine only the recursive child evidence.

\tcblower
\textbf{Per-position input template}

\begin{flushleft}
\ttfamily\footnotesize
ROOT FEN: <root FEN>\\
ROOT SIDE TO MOVE: <White or Black>\\[0.5em]
MANDATORY BEST ROOT CANDIDATE: <tokenized move> (<UCI move>)\\[0.5em]
CHILD X\\
ROOT CANDIDATE: <tokenized move> (<UCI move>)\\
CHILD FEN: <child FEN>\\
MODEL-IMPLIED ROOT WIN RATE: <percentage>\%\\
<critical-line boundary status>\\
CHILD ANALYSIS (opponent POV; move numbering refreshed):\\
<child analysis and any applicable mate-distance note>\\[0.5em]
<corresponding blocks for CHILD Y and CHILD Z>
\end{flushleft}
\end{tcolorbox}
Finally, below is the prompt we give to the GPT-5.6-Sol (high) judge when scoring explanations on the coherence and fluency axes:

\begin{tcblisting}{
  title={GPT-5.6-Sol (high) evaluation prompt},
  breakable,
  colback=white,
  colframe=black!60,
  listing only,
  listing options={
      basicstyle=\ttfamily\footnotesize,
      breaklines=true,
      columns=fullflexible,
      keepspaces=true,
      showstringspaces=false
  }
}
Position FEN: {{FEN}}
Side to move: {{SIDE_TO_MOVE}}

Stockfish reference (1,000,000 nodes per root search; every evaluation is from the
perspective of the side to move in the root position):
1. {{STOCKFISH_PV_1_MOVE}} - evaluation {{STOCKFISH_PV_1_EVAL}}; PV:
{{STOCKFISH_PV_1}}
2. {{STOCKFISH_PV_2_MOVE}} - evaluation {{STOCKFISH_PV_2_EVAL}}; PV:
{{STOCKFISH_PV_2}}
3. {{STOCKFISH_PV_3_MOVE}} - evaluation {{STOCKFISH_PV_3_EVAL}}; PV:
{{STOCKFISH_PV_3}}

Act as a chess grandmaster and rate the following analysis of the above position on
the following three axes:
(i) Lack of *structural* hallucinations
(ii) Lack of *conceptual* hallucinations
(iii) Fluency
You will rate each factor on a scale of 1-5. The information below provides more
detailed instructions on the interpretation of each factor and its associated score.

**Structural hallucinations**
Structural hallucinations are the hallucinations related to the moves and
variations discussed in the response, as well as in the model's final critical
variation. In particular, considering mistakes or illegal moves in one's analysis
leads to a lower score on this axis. Note, of course, that considering mistakes
for illustrative purposes and discussing why they are bad alternative moves is acceptable. In general, the severity of a penalty is directly affected
by how severely the mistake or illegal move affects the overall analysis and
conclusion. In addition, note that mistakes and illegal moves later into the
variation are usually less severe than those earlier. Furthermore, illegal moves
are worse than mistakes. For this score, you should ignore any conceptual
hallucinations in the response.

The final critical line is a crucial component of the structural score because
it directly expresses the conclusion of the analysis. Give particular weight
to mistakes or illegal moves there, especially near the beginning. Use the
automatic per-move blundercheck and its win-rate drops as evidence, while still
judging the response as a whole.

The interpretation of the various scores here are:
1: The entire response consists of hallucinated moves, and no proper move or
variation may be understood from it.
2: The response consists majorly of mistaken or illegal moves, which substantially
affect the takeaways. There do exist some legal or correct moves considered, but
they are few.
3: The response is generally correct, but has seriously blundered or illegal moves
which directly affect a verdict of whether a certain move is playable/best.
Nonetheless, the general analysis holds.
4: The response is mostly correct. There is at least one mistaken detail, but the
few that exist are deep into variations and/or do not substantially affect the
evaluation of the move(s) considered close to the root position.
5: Perfect response: there are no hallucinations in terms of blunders or illegal
moves.

**Conceptual hallucinations**
Conceptual hallucinations are the hallucinations related to the patterns and themes
of the position at hand. For instance, calling a move a fork when it does not
attack two pieces, or claiming that white has a lead in development when they do
not, would be classified as a conceptual hallucination. Once again, the penalty of
a conceptual hallucination is tied to how severly it affects the coherence of the
analysis. For instance: calling a strongly placed knight an outpost may be a minor
infraction as long as the knight is indeed strong and well-supported on the square,
and the subsequent justification is unaffected by it. For this score, you should
ignore any structural hallucinations in the response.

The interpretation of the various scores here are:
1: The entire justification consists of fabricated themes and motifs, almost none
of which are materially true.
2: The response consists majorly of conceptual hallucinations. These substantially
degrade the intelligibility of the response, though at least a rough sense of the
position can be inferred.
3: The response is generally correct, but has serious hallucinations of motifs or
patterns. These do affect the readability of the response, but still do not overall
change the verdict and evaluation of the position.
4: The response is mostly correct. There are a few hallucinated terms or motifs,
but the justification and intent of the response can be largely inferred without
confusion.
5: Perfect response: there are no conceptual hallucinations and the depicted themes
are correct and coherent.

**Fluency**
Fluency measures the command of the response over the English language: whether,
first of all, it attaches sufficient justification to its moves (e.g., explains why
a certain alternative is worse, rather than just throwing three moves at the
reader); secondly, whether it uses clearly legible and grammatical sentences to
describe the position's analysis. Please refer to the note below as well. The
interpretation of various scores are:
1: The response only provides the move or few moves, and no justification is given
for them.
2: The response provides a very terse justification for its moves which do not
explain them or consider alternatives.
3: The response provides some justification, but the prose is awkward or
ungrammatical.
4: The response is largely legible. Some words or phrases are not entirely clear in
their meaning.
5: The prose in the response can be completely comprehended by a typical chess
player fluent in English.

Note about some of the explanations: Some explanations may include tokens like
<WHITE_ROOK>, <SQUARE_A5>. These are piece tokens output by the chess-playing model
and as such should not be penalized in any of the above metrics. Moreover, since
the model may view the board from the to-move POV (reflected), it may discuss white
moves with 1... instead of 1; or it may call dark squares as light squares (and
vice versa). These two quirks are also not to be penalized.

With the above instruction, here is the response to be judged:

>>>RESPONSE>>>
{{RESPONSE}}
<<<RESPONSE<<<
Automatic blundercheck of critical line:
{{CRITICAL_LINE_BLUNDERCHECK}}

Stockfish evaluation after the first move in the critical line, from the original
side-to-move perspective: {{CRITICAL_LINE_FIRST_MOVE_EVAL}}

Return only one JSON object in exactly this form:
{
"lack_of_structural_hallucinations": {
  "score": <integer from 1 to 5>,
  "reason": "<brief explanation citing the most important concrete structural
  evidence>"
},
"lack_of_conceptual_hallucinations": {
  "score": <integer from 1 to 5>,
  "reason": "<brief explanation citing the most important concrete conceptual
  evidence>"
},
"fluency": {
  "score": <integer from 1 to 5>,
  "reason": "<brief explanation citing the most important evidence about
  readability and justification>"
}
}

All three scores are required and must be integers. Judge the three axes
independently. Do not include an overall score, Markdown, or any text outside the
JSON object.
\end{tcblisting}

\end{document}